\pdfoutput=1
\documentclass[11pt]{article}

\usepackage[final]{coling}

\usepackage{times}
\usepackage{latexsym}
\usepackage{array}
\usepackage{graphicx}
\usepackage{epstopdf}
\usepackage{makecell}
\newcolumntype{L}[1]{>{\raggedright\let\newline\\\arraybackslash\hspace{0pt}}m{#1}}
\newcolumntype{C}[1]{>{\centering\let\newline\\\arraybackslash\hspace{0pt}}m{#1}}
\newcolumntype{R}[1]{>{\raggedleft\let\newline\\\arraybackslash\hspace{0pt}}m{#1}}
\usepackage{multirow}
\usepackage{svg}
\usepackage{makecell}
\usepackage[utf8]{inputenc}
\usepackage{amsmath}
\usepackage{cleveref}
\crefname{section}{§}{§§}
\Crefname{section}{§}{§§}
\usepackage{booktabs}
\usepackage{xcolor} 
\usepackage{colortbl}
\usepackage{color}
\usepackage{lipsum}
\usepackage[T1]{fontenc}
\usepackage[utf8]{inputenc}

\usepackage{microtype}

\usepackage{inconsolata}

\usepackage{graphicx}

\title{Awakening Augmented Generation: Learning to Awaken Internal Knowledge of Large Language Models for Question Answering}

\author{Huanxuan Liao$^{1,2}$, Shizhu He$^{1,2}$\thanks{Corresponding author}, Yao Xu$^{1,2}$, Yuanzhe Zhang$^{1}$,\\ \textbf{Shengping Liu}$^{3}$, \textbf{Kang Liu}$^{1,2}$, \textbf{Jun Zhao}$^{1,2}$  \\
    $^1$ The Key Laboratory of Cognition and Decision Intelligence for Complex Systems, \\
    Institute of Automation, Chinese Academy of Sciences, Beijing, China \\
    $^2$ School of Artificial Intelligence, University of Chinese Academy of Sciences, Beijing, China \\
    $^3$ Unisound, Beijing, China\\
  {liaohuanxuan2023@ia.ac.cn} {\{yao.xu, shizhu.he, kliu, jzhao\}@nlpr.ia.ac.cn} \\}

\begin{document}
\maketitle
\begin{abstract}
Retrieval-Augmented-Generation and Genera-tion-Augmented-Generation have been proposed to enhance the knowledge required for question answering with Large Language Models (LLMs) by leveraging richer context. However, the former relies on external resources, and both require incorporating explicit documents into the context, which increases execution costs and susceptibility to noise data during inference. Recent works indicate that LLMs model rich knowledge, but it is often not effectively activated and awakened. Inspired by this, we propose a novel knowledge-augmented framework, \textbf{Awakening-Augmented-Generation} (AAG), which mimics the human ability to answer questions using only thinking and recalling to compensate for knowledge gaps, thereby awaking relevant knowledge in LLMs without relying on external resources. 
AAG consists of two key components for awakening richer context. Explicit awakening fine-tunes a context generator to create a synthetic, compressed document that functions as symbolic context. Implicit awakening utilizes a hypernetwork to generate adapters based on the question and synthetic document, which are inserted into LLMs to serve as parameter context. Experimental results on three datasets demonstrate that AAG exhibits significant advantages in both open-domain and closed-book settings, as well as in out-of-distribution generalization. Our code will be available at \url{https://github.com/Xnhyacinth/IAG}.
 %\blfootnote{* Corresponding Author}
\end{abstract}

\section{Introduction}
% \colorbox {gray!20}{\textit{We can know more than we can tell. \\--- Michael Polanyi}}
\textcolor[RGB]{161, 64, 0}{\textbf{\textit{We can know more than we can tell. --- Michael Polanyi}}}
% \textcolor{brown!150}{\textbf{\textit{We can know more than we can tell. --- Michael Polanyi}}}
\vspace{0.2cm}
% \hl{\textit{We can know more than we can tell. \\--- Michael Polanyi}}
% \textcolor[RGB]{161, 64, 0}{\textbf{\textit{We can know more than we can tell. --- Michael Polanyi}}}

Knowledge-intensive tasks like question answering (QA) necessitate utilizing extensive world and domain knowledge \cite{wq, tqa, nq}.
% , rendering these tasks challenging even for humans without an external knowledge source like Wikipedia. 
Nowadays, Large Language Models (LLMs) have displayed notable competencies in almost every task and industry \cite{pre}.
%within the “pre-train, prompt, and predict” paradigm 
However, LLMs lack the sufficient capability to independently handle knowledge-intensive tasks \cite{frisoni2024generate} and usually generate hallucinations \cite{knowing}.

\begin{figure}[t]
\centerline{\includegraphics[width=0.5\textwidth]{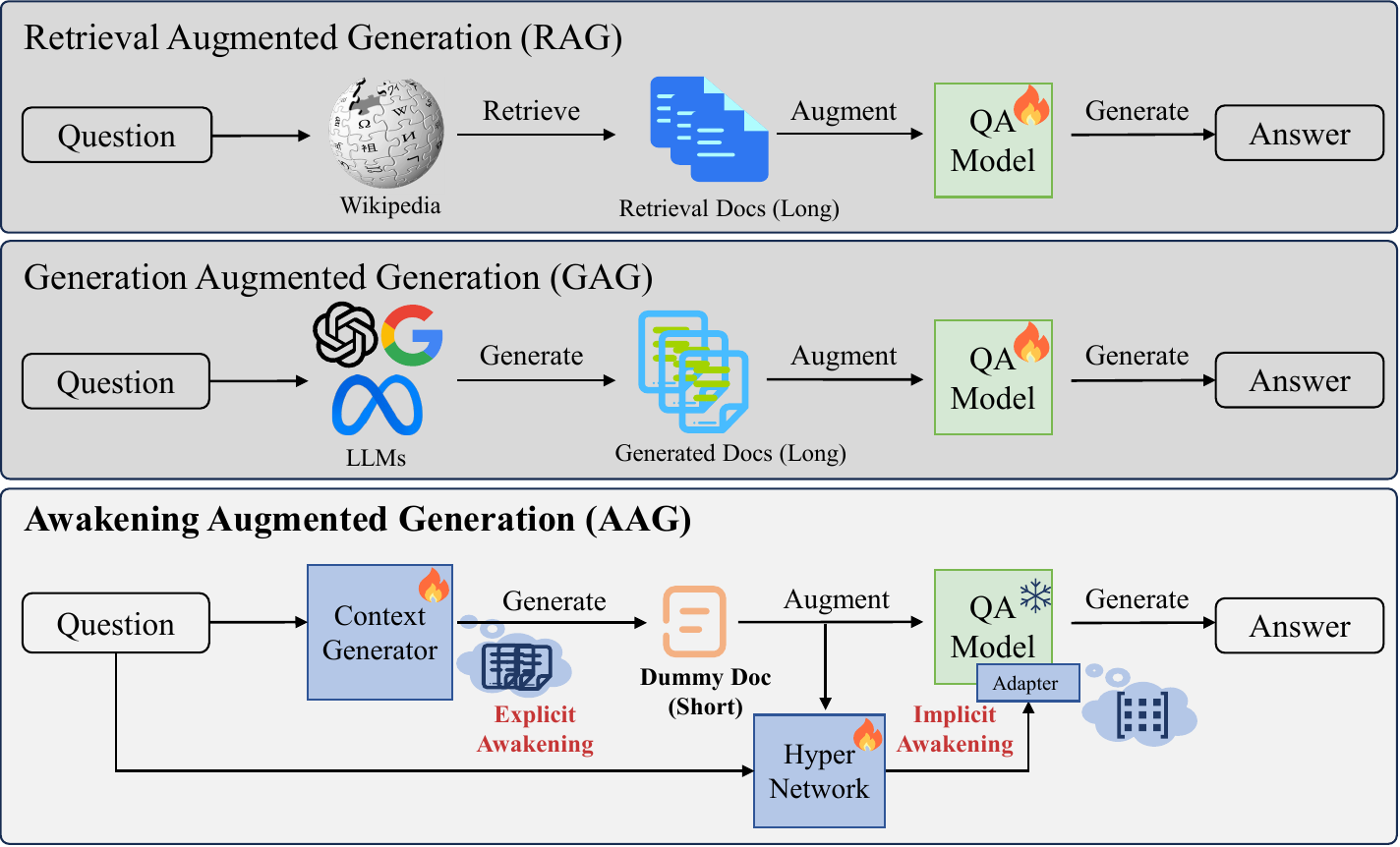}}
% \vspace{-0.45cm}
\caption{Compared with RAG  and GAG, the proposed AAG  eschews external resources, generates a dummy document (explicit awakening) and creates flexible adapters (implicit awakening) for each question.
}
\label{intro}
% \vspace{-0.65cm}
\end{figure}

In recent years, to address hallucinations in LLMs and enhance performance in question answering, researchers have developed several knowledge-augmented methods for LLMs. These methods primarily fall into two categories: Retrieval-Augmented Generation (RAG) \cite{REALM} which retrieves documents from external resources (e.g., Wikipedia) and incorporates both the retrieved documents and the question into LLMs \cite{fid} (top part of Figure~\ref{intro}). Generation-Augmented Generation (GAG) \cite{sure} which utilizes LLMs such as ChatGPT \cite{training} to generate more relevant documents, which are then used to enhance the answer generation (middle part of Figure \ref{intro}).
% Retrieval-Augmented-Generation (RAG) \cite{REALM} and Generation-Augmented-Generation (GAG) \cite{sure}.
% RAG (Top part of Figure~\ref{intro}) retrieves related documents from external resources (e.g., Wikipedia) and then inputs both the retrieved documents and the question into LLMs \cite{fid}. GAG (Middle part of Figure \ref{intro}) leverages external LLMs such as ChatGPT \cite{training} to generate relevant documents \cite{generate}.

However, these methods have the following disadvantages\footnote{A more intuitive comparison can be seen in \ref{method_com}.}:
1) \textbf{Dependence on external resources}, RAG relies on external domain knowledge resources \cite{bridging}, while GAG depends on a more powerful external LLM as a knowledge generator. This reliance limits their broader application.
2) \textbf{Increased execution costs}, the computing resources and inference time required increase significantly with the number of documents. For example, the typical RAG method FiD \cite{fid} must handle over 12,000 tokens to retrieve 100 documents, resulting in more than a $100$-fold increase in prompt length and over $100^2$-fold increase in inference time \cite{lost}. Similarly, the GAG method \cite{generate} incurs additional financial costs, such as API calls.
3) \textbf{Specific retraining}, these approaches often require retraining for different domains, tasks and datasets \cite{mend}. This heightens the challenge of reusing models across different scenarios, resulting in resource inefficiency due to low parameter effectiveness and the need for extensive data.
% In contrast, hypernetworks \cite{hypernetworks} offer a solution that reduces the dependency on gradient descent for specific domains. Methods such as Hypertuning \cite{hypertuning} and HINT \cite{hint} use hypernetworks to transform inputs into parameter-efficient modules, thereby reducing computation and enhancing model generalization.
% 3) \textbf{Susceptibility to noise data}, both methods are prone to the inclusion of irrelevant or noisy data \cite{compressing, desiderata}. For example, \citet{retrievalaugmented} indicates that noise in the documents negatively affects the performance. Therefore, there is an urgent need to explore new knowledge-augmented methods.

In fact, LLMs inherently possess rich knowledge and significant potential for tackling knowledge-intensive tasks \cite{abductive}. Performance on specific tasks can be improved by more effectively activating and awakening relevant knowledge without external resources.
% or by expanding memory capacity without relying on external resources.
For instance, strategies such as repeating the question twice \cite{rereading}, consolidating knowledge with prompts like "\textit{As far as I know}" \cite{knowledger}, and employing visual-language models to imagine images \cite{learning-imagine} can all enhance the performance of LLMs on downstream tasks. That is, \textbf{LLMs model rich knowledge, but it is often not effectively activated and awakened}. 

Inspired by the above findings and to alleviate the challenges in RAG and GAG, we propose a novel knowledge-augmented framework called \textbf{A}wakening-\textbf{A}ugmented \textbf{G}eneration (AAG) which emulates the human ability to compensate for knowledge deficits through thinking and recalling in QA. 
AAG utilizes the context generator to generate a compressed dummy document as symbolic context while reducing computational demands. For instance, AAG uses "\textit{official language ... Jamaica}" (just 20 tokens) as knowledge instead of "\textit{Jamaica is regarded... official language is English...}" (>200 tokens) in RAG or GAG for the question "\textit{what does jamaican people speak?}" in WebQ \cite{wq}. Additionally, AAG uses the hypernetwork to generate adapters as parameter context for each question, which integrates the advantages of instruction-based learning with parameter-efficient modules to awaken a richer context in LLMs (bottom part of Figure \ref{intro}).
% Therefore, inspired by the above phenomena and aiming to address the challenges faced by RAG and GAG, we propose a novel knowledge-augmented framework, Awakening-Augmented-Generation (AAG) which simulates the human capacity to compensate for knowledge deficits solely through imagination in QA. For resolving knowledge-intensive tasks,  AAG utilizes LLMs to generate concise contexts and design adaptable modules, thereby refining, enhancing accuracy, and ensuring the relevance of input text. This approach reduces computational demands, leading to cost savings and faster inference. Concurrently, AAG employs hypernetworks to produce question-specific parameters tailored to efficient modules based on the input, effectively integrating the merits of instruction-based learning with parameter-efficient modules (Bottom part of Figure \ref{intro}). 
% Through a knowledge distillation framework, AAG achieves knowledge enhancement through imagination. 
% The

% Its core idea is to \textbf{enable student models, which lack rich contextual information, to mimic teacher models that possess such information}.

% Within the framework of AAG, we introduce \textbf{im}agine richer \textbf{c}ontext method for \textbf{q}uestion \textbf{a}nswering (AAG). 
Specifically, to sufficiently awaken the inherent knowledge of LLMs, we design two main modules to obtain different types of contexts and improve the utilization of relevant knowledge in LLM. The \textbf{explicit awakening} module first employs symbol distillation to compress context, followed by fine-tuning the context generator to generate a concise dummy document, effectively reducing the length of text processing. Next, within the knowledge distillation framework, the \textbf{implicit awakening} module utilizes a hypernetwork to convert questions and other task data (e.g., documents) into adapters inserted into LLMs. This dynamic generation allows for more adaptable and contextually relevant module generation, enhancing the model's ability to handle diverse and complex tasks effectively. The core idea of AAG is to \textbf{enable student models that lack rich contextual information to mimic teacher models that possess such information}.

We evaluate the proposed AAG on various LLMs, including T5 \cite{T5} and Llama2 \cite{llama}. The experimental results across NQ \cite{nq}, TriviaQA \cite{tqa} and WebQ datasets indicate that the proposed AAG yields performance gains while reducing computational expenses and time during inference. Notably, it outperforms baselines that retrieve and generate knowledge $2\%$ under the same document settings and can achieve similar performance while reducing inference cost (tokens processed) by up to $4\times$. In conclusion, the contributions of this paper are summarized as follows:
\begin{itemize}
    \item We propose a new knowledge augmentation framework AAG to awaken richer context (symbolic and parameter context) more efficiently without relying on external resources.
    % \item We propose a novel QA method AAG that employs two main modules (explicit awakening and implicit awakening) to utilize better the knowledge stored in the LLMs and awaken a richer context in QA. 
    \item  We make use of a text-conditioned hypernetwork to generate parameter-efficient modules as parameter context based on the question and a dummy compressed document.
    \item Experimental results indicate that AAG effectively awakens the relevant knowledge of LLMs which demonstrates significant advantages in both open-domain and closed-book settings while reducing inference cost.
    % Additionally, it excels in out-of-distribution generalizations. 
\end{itemize}

\section{Related Work}

This paper mainly utilizes context compression, hypernetworks and knowledge distillation to achieve knowledge enhancement. The following will elucidate pertinent research across four facets.

\noindent \textbf{Knowledge Enhancement} has usually been adopted to alleviate the issue of insufficient knowledge in LLMs. There are two main methods: RAG \cite{pullnet, hintenhanced} and GAG \cite{generator}.
The typical RAG method FiD~\cite{fid} retrieves documents from Wikipedia to answer questions. LLMs serving as a knowledge base have been the focus of numerous studies that advocate the extraction of knowledge from such models (e.g., GPT-3). For instance, \citet{generate} generates 10 documents for each question. However, RAG requires external resources, and both RAG and GAG need verbose long contexts. Recently, methods have been developed to enhance LLMs' abilities by simulating human imagination of visual information using existing visual-language models \cite{learning-imagine, selfimagine}. 
% We prefer self-imagination to augment knowledge and aim to leverage the parameterized knowledge within the models \cite{rereading, measuring}. 
Our proposed method not only eliminates the need for external resources but also improves the efficiency of activating internal knowledge within LLMs.
% The typical RAG method FiD~\cite{fid} retrieves the documents from Wikipedia to answer questions. As LLMs can be considered a knowledge base, several studies \cite{generated} propose to extract knowledge from LLMs (e.g., GPT-3). For example, \citet{generate} generates 10 relevant documents for world knowledge according to the question. However, RAG needs the related external resources, and both RAG and GAG still need to obtain and utilize verbose explicit long contexts. Recently, there have been some methods to enhance the LLMs' ability through simulating human imagination of visual information. But they use existing visual-language models \cite{learning-imagine, selfimagine}, while we prefer self-imagination to augment knowledge. Besides, they have not fully leveraged the parameterized knowledge within the models \cite{rereading, measuring}. In this paper, the proposed method for augmenting knowledge not only obviates the need for external resources but also enhances the efficiency of extracting and activating internal knowledge within LLMs.

% \vspace{0.1cm}
\noindent \textbf{Context Compression} has often been used to improve the efficiency of LLMs in processing long contexts. Recent studies \cite{learning} propose that long contexts be condensed into summary vectors (soft prompts) to ensure their effective utilization by LLMs. %Such soft prompts can efficiently replace plain text demonstrations, consequently reducing computational expenses during the inference stage. 
Simultaneously, some studies \cite{longllmlingua, llmlingua2} suggest utilizing information redundancy and entropy in lengthy texts to compress contexts \cite{compressing}. Unlike these approaches, this paper aims to enhance the long-context modeling ability of LLMs. By developing a context generator that creates compressed contexts, the QA model operating on short contexts can achieve a rich contextual understanding similar to models designed for longer contexts.
% Simultaneously, some studies \cite{longllmlingua} believe that information redundancy in lengthy texts and information entropy can be utilized to compress the contexts \cite{compressing}. Unlike them, this paper is devoted to awakening the long-context modeling ability of LLMs. By learning a context generator that can generate compressed contexts, the QA model that operates on short contexts can also possess a rich contextual understanding akin to the QA model designed for processing longer contexts.

% \vspace{0.1cm}
\noindent \textbf{Knowledge Distillation} is a technique where a smaller model learns to mimic the predictions of a larger model, aiming to retain performance while reducing computational resources \cite{kd}. Recent studies \cite{symbolic} present symbolic knowledge distillation, a process that facilitates knowledge transfer from a teacher model via extracting training data to subsequently train a student model \cite{scott, aligning}. In this paper, the process of obtaining compressed context during context generator fine-tuning resembles a form of symbolic distillation. Regarding training, our emphasis lies in distilling the long-context modeling abilities of LLMs. %Both teacher and student models possess identical sizes (comprising the same layers and heads), albeit with varying context lengths.

% \vspace{0.1cm}
\noindent \textbf{Hypernetworks} is designed to reduce the number of parameters \cite{hypernetworks}, i.e., a small neural network generates parameters for another big neural network. It offers a solution that reduces the dependency on gradient descent for specific domains. Recent studies \cite{hypertuning, hint} have explored the enhancement of model performance in zero/few-shot settings through meta-learning involving hypernetworks.
% to reduce computation and enhance model generalization. 
We utilize hypernetworks to acquire parameter context by dynamically converting the question and the other data to adapters inserted into LLMs for efficiency and generalization.

\section{Method}
In this section, we introduce the details of AAG to activate LLMs' intrinsic knowledge and obtain a richer context for QA. The fundamental premise underlying this method is that QA with a richer context (teacher model) yields a better internal representation and greater performance (e.g., RAG with retrieved documents). Therefore, to enable a student model without external documents as context to also possess rich context, it is necessary to both learn to independently generate context (though not excessively long) and to allow the student model to mimic and acquire rich internal representations.
% In this section, we introduce the details of AAG to activate LLMs' intrinsic knowledge and obtain a richer context for QA. The fundamental premise underlying this method is that QA with a richer context yields greater performance (e.g., RAG with retrieved documents). Consequently, diverse methods are employed for questions lacking in richer contexts to activate knowledge within LLMs to replicate comparable effects to those achieved with richer contexts.

\begin{figure*}[t]
\centerline{\includegraphics[width=\textwidth]{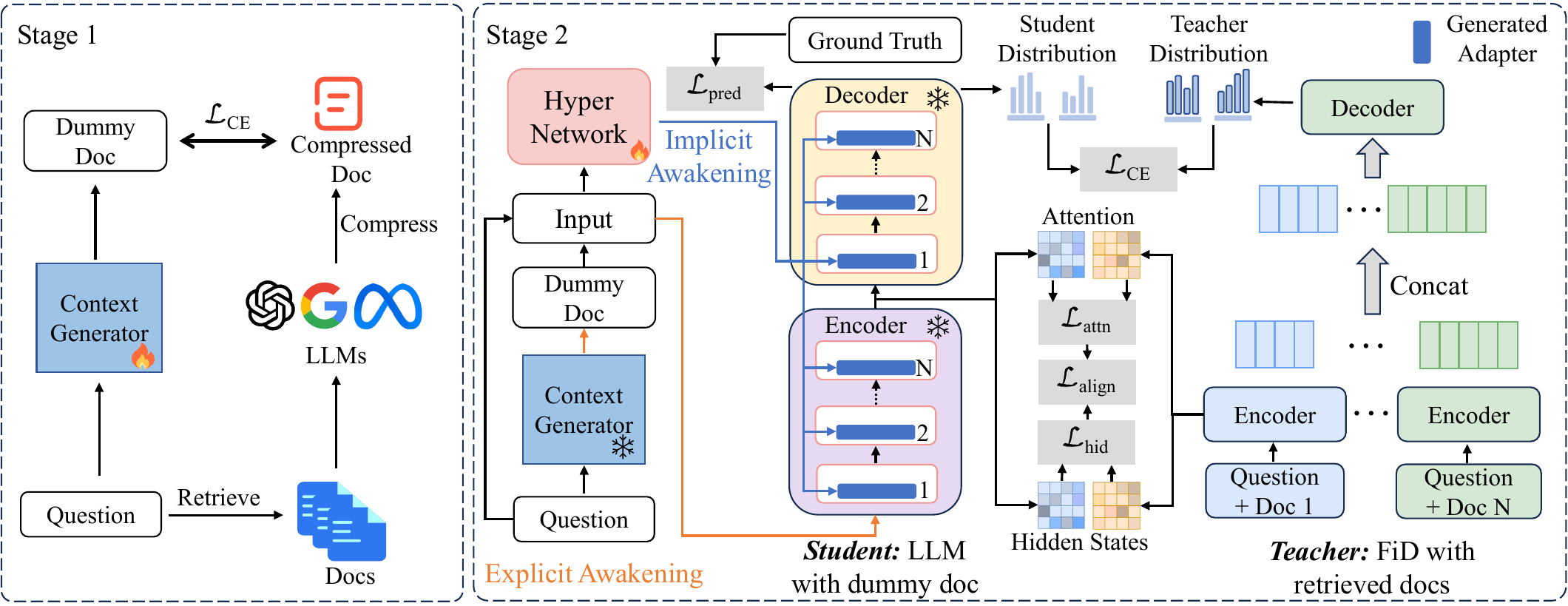}}
% \vspace{-0.45cm}
\caption{Overview of AAG method. In the inference phase, for each question, the explicit awakening (context generator) generates a short dummy document and the implicit awakening (hypernetwork) generates a specific LoRA module. During training, there are two stages: the first stage is the \textbf{pre-training} of the \textbf{context generator} (\cref{ssec:Compress for Explicit Knowledge}), aiming at its ability to imagine a short dummy document based on the question, and the second stage is the \textbf{hypernetwork fine-tuning} (\cref{ssec:Hypernetwork}) using long context distillation (\cref{ssec:Knowledge Distillation for contextualized Knowledge}) to obtain a question-specific LoRA module.}
% the compression of retrieved documents into condensed text, which supervises the \textbf{context generator} to imagine a dummy document and the Hypernetwork learns a map from the question to the LoRA weights. Besides, the teacher model with richer context aligns hidden representations, imagining additional knowledge.
\label{fig1}
% \vspace{-0.65cm}
\end{figure*}

Specifically, as shown in Figure~\ref{fig1}, AAG comprises two main modules. \textbf{Explicit awakening} with long context compression learns to generate a compressed dummy document (\cref{ssec:Compress for Explicit Knowledge}). \textbf{Implicit awakening} with the hypernetwork leverages hidden knowledge that learns a shared knowledge feature projection across questions (\cref{ssec:Hypernetwork}). The hypernetwork is trained to generate lightweight LoRA modules to align the question and the internal knowledge. Besides, there is long context distillation in training, which learns the teacher's rich representations to compensate for missing knowledge in label learning (\cref{ssec:Knowledge Distillation for contextualized Knowledge}).

\subsection{Formulation}

The formulation of our task follows RAG for QA \cite{REALM}. Let $\mathcal{V}^*$ denote the infinite set, encompassing all potential strings over the tokens in vocabulary $\mathcal{V}$, and this includes the empty string. An instance within a QA dataset is defined as a triplet $(\boldsymbol{q}, \boldsymbol{a}, \boldsymbol{c})$ comprising question $\boldsymbol{q}$, answer $\boldsymbol{a}$, and context $c$, where $\boldsymbol{q}, \boldsymbol{a}, \boldsymbol{c} \in \mathcal{V}^*$. Conventionally, the context $c$ is drawn from the knowledge corpus $\mathcal{Z}$, like Wikipedia, whereby $\mathcal{Z} \subset \mathcal{V}^*$. Additional background details are available in \ref{sec:back}.

% The goal of QA is to learn a distribution function \( p(\boldsymbol{a}|\boldsymbol{q}) \). In a closed-book setting, LLMs directly encode the given question $\boldsymbol{q}$ and generate the answer $\boldsymbol{a}$ \cite{much}. However, employing a direct approach of requesting models to output answers frequently results in poor performance, primarily attributable to omitting a substantial amount of world knowledge. Therefore, a popular approach is the open domain setting, which marginalizes \( p(\boldsymbol{a}|\boldsymbol{q}, \boldsymbol{c}) \) over contexts $c$. 

\subsection{Explicit Awakening with Context Generator}
\label{ssec:Compress for Explicit Knowledge}

To obtain the short dummy document $\boldsymbol{d}$, we fine-tune a context generator \footnote{We discuss the role of context generator in the \ref{role}.} to utilize its knowledge in generating a compressed dummy document as symbolic context, thereby reducing input length. Simultaneously, we avoid dependence on a fixed knowledge base and minimize \textit{knowledge corpus errors} by incorporating potentially useful context \cite{knowledge}. Employing a knowledge distillation framework, the student model learns to generate the compressed text that the teacher model produces based on extensive context.

Specifically, for each data point $\mathcal{D}_\text{train} = \{(\boldsymbol{q}_i, \boldsymbol{a}_i, \boldsymbol{c}_i)\}_{i=1}^n$, we apply the long-context compression method LongLLMLingua \cite{longllmlingua} to the retrieved text $\boldsymbol{c}_i$, resulting in the compressed text $\boldsymbol{c'}_i$. As shown in the left part of Figure \ref{fig1}, subsequently, we fine-tune the context generator $p_\theta$ with trainable parameters $\theta$ to fully leverage its inherent knowledge for generating $\boldsymbol{c'}_i$, which guides the model to think about its knowledge and generate a short dummy document. Our objective is to minimize the negative log-likelihood of the compressed text $\boldsymbol{c'}_i$ sequence given the specific prompt $\boldsymbol{p}$ (\ref{sec:prompts}) and the question $\boldsymbol{q}_i$.
% As shown in the left part of Figure \ref{fig1}, to help LLMs fully utilize the knowledge and generate compressed text, we first pretrain the context generator on our collected question-compressed document pairs. By leveraging symbolic distillation, we employ the long context compression method LongLLMLingua \cite{longllmlingua} to condense a large corpus of retrieved documents. These compressed texts $c'$ then serve as fine-tuning data alongside specific prompts $p_q$ (\ref{sec:prompts}) and question-answer pairs for the context generator\footnote{We discuss the role of context generator in the \ref{role}.} $p_\theta$ ($\theta$ represents the model’s parameter), which guides the model to think about its knowledge and generate a short dummy document:
\begin{equation}
    % d = G_\theta(p_q(q))
    \mathcal{L}_{ce} =  - \frac{1}{n} \sum_{i=1}^{n} \log p_\theta(\boldsymbol{c'}_i \mid \boldsymbol{p}, \boldsymbol{q}_i)
\end{equation}
This process enables LLMs to conceive compressed document that robustly parallels the question's knowledge requirements.

\begin{figure}[t]
\centerline{\includegraphics[width=0.5\textwidth]{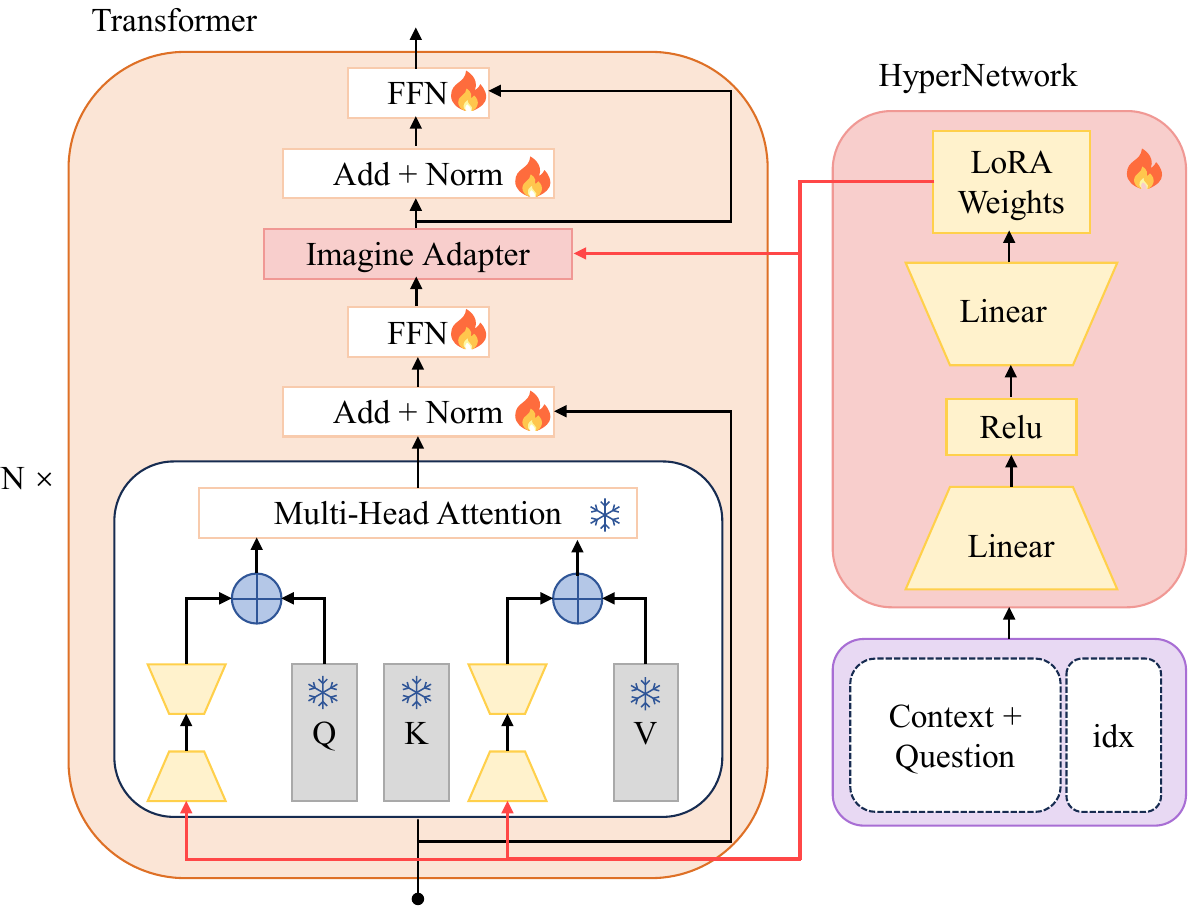}}
% \vspace{-0.45cm}
\caption{The Architecture of hypernetwork. Hypernetwork generates LoRA adapter weights for each question. During training, only Hypernetwork, FFN, and Norm weights are updated.}
\label{fig2}
% \vspace{-0.65cm}
\end{figure}

\subsection{Implicit Awakening with Hypernetwork}
\label{ssec:Hypernetwork}
Generally speaking, richer context can help LLM better answer questions. That is, the representation of questions and the internal state of LLM when utilizing rich context are the better states. Therefore, in the absence of context, we should focus on building models to awaken LLM to achieve this better state and as a better QA model.

We utilize the hypernetwork\footnote{We conduct a detailed analysis of the reasons behind the hypernetwork in the \ref{reason}.} to convert the question $\boldsymbol{q}$ and short dummy document $\boldsymbol{d}$ into a specific parameter-efficient LoRA module inserted into the LLM, serving as the parameter context for the question.
% does not directly optimize the LoRA \cite{LoRA} module but generates specific LoRA adapter weights using the inputs for QA (bottom part of Figure \ref{fig1}). 
This is akin to repeating the question in the prompt \cite{rereading} and incorporating certain topical cues to stimulate the model's recall of relevant questions \cite{large}. However, the distinction lies in the fact that they serve as wake-up features, whereas we are generating model parameters as knowledge awakening. 

% The hypernetwork architecture for generating LoRA weights is exhaustively outlined in Figure \ref{fig2}. Specifically, $D_k^q$ and $U_k^q$ represent the low-rank projections of layer $k$ correlated with the $\mathcal{Q}$, while $D_k^v$ and $U_k^v$ correspond to those associated with the $\mathcal{V}$. Hypernetwork represented as $g_D$ and $g_U$, takes $\textit{concat}(f, i_k^{\{q,v\}})$ as input, where $f$ is the feature vectors that use the model's encoder to obtain and using whitening algorithm \cite{bert} for dimensionality reduction, $i_k^{\{q,v\}} \in \{0, \ldots, 2 \times \# \text{blocks}\}$ signifies the positional embedding differentiating between layers and between $\mathcal{QV}$. Each hypernetwork is defined by weights $W_d$ and $W_u$ which represent the down and up projections respectively. Finally, the hypernetwork equations for $D^{\{q,v\}}$ and $U^{\{q,v\}}$ can be expressed as:
The hypernetwork architecture for generating LoRA weights is detailed in Figure \ref{fig2}. Specifically, $D_k^q$ and $U_k^q$ represent the low-rank down and up projections of layer $k$ associated with the \textit{Query} matrix $W_\mathcal{Q}$ in the attention module, while $D_k^v$ and $U_k^v$ correspond to those associated with the \textit{Value} matrix $W_\mathcal{V}$. The hypernetwork, denoted as $g_D$ and $g_U$, takes $\textit{concat}(f, i_k^{\{q,v\}})$ as input, where $f$ is the feature vector obtained using the model's encoder and reduced in dimensionality via a whitening algorithm \cite{bert}. To achieve this whitening transformation, we first compute the mean of the vector \( \mu = \frac{1}{N} \sum_{i=1}^N x_i \) and center the data by subtracting \( \mu \) from each vector \( x_i \). Next, we calculate the covariance matrix \( C \) of the centered vectors \( \tilde{x}_i = x_i - \mu \), which is given by \( C = \frac{1}{N} \sum_{i=1}^N \tilde{x}_i \tilde{x}_i^T \). We then perform Singular Value Decomposition (SVD) on the covariance matrix: \( C = U \Lambda U^T \), where \( U \) contains the eigenvectors and \( \Lambda \) is a diagonal matrix of eigenvalues. The transformation matrix \( W \) is derived from the eigenvalue decomposition as \( W = U \Lambda^{-1/2} \), where \( \Lambda^{-1/2} \) scales the eigenvectors by the inverse square root of their corresponding eigenvalues. Thus, applying the transformation \( \tilde{x}_i = (\tilde{x}_i) W \) not only centers the data around zero but also results in a covariance matrix that is equivalent to the identity matrix, ensuring that the transformed vectors are uncorrelated and have unit variance. The term $idx_k^{\{q,v\}} \in \{0, \ldots, 2 \times \# \text{blocks}\}$ signifies the positional embedding, differentiating between layers and $\mathcal{QV}$. Each hypernetwork is characterized by weights $W_d$ and $W_u$, representing the down and up projections, respectively. The hypernetwork equations for $D^{\{q,v\}}$ is expressed as follows:
\begin{equation}
   f_i = \text{whitening}(\text{Encoder}(\boldsymbol{q}_i;\boldsymbol{d}_i))
\end{equation}
% \begin{equation}
%     x_i = \text{concat}(f_i, idx_k^{\{q,v\}} ~|~ i_k^q = 2k, i_k^v = 2k+1)
% \end{equation}
\begin{equation}
% g(x) = \text{MM}(\text{ReLU}(\text{MM}(x, W_d)), W_u)
g(x) = W_u \cdot \text{ReLU}(W_d \cdot x)
\end{equation}
\begin{equation}
    % D^{\{q,v\}}, U^{\{q,v\}} = g_D(x), g_U(x)
    D^{\{q,v\}} = g_D((f_i; idx_k^{\{q,v\}}))
\end{equation}
where Encoder represents the encoder of the model, whitening is a dimensionality reduction algorithm, $\text{ReLU}$ is an activation function, and $idx_k^q = 2k, idx_k^v = 2k+1$. $g_D$ and $g_U$ represent the dimension reduction and dimension increase functions of the hypernetwork, respectively.

\subsection{Training with Long Context Distillation}
\label{ssec:Knowledge Distillation for contextualized Knowledge}

% Within the framework of knowledge distillation, components such as hidden representations \cite{tinybert}, attention dependencies \cite{minilm}, and relations among representations \cite{distilling} are regarded as valuable knowledge for transfer. In this paper, we consider long context distillation (LCD) as the contextualized knowledge that mainly guides the student.

% Specifically, the teacher model FiD \cite{fid}, which utilizes longer contextual inputs and theoretically contains more information (richer context). It will activate more specific internal knowledge and serve as a supervisory model. The teacher model assists the student model T5 \cite{T5}, which has the same size as the teacher and leverages short contextual inputs. This aids in activating richer feature representations and knowledge. %, while also instructing it on activating corresponding internal knowledge and imagining related information. 
Within the knowledge distillation framework, elements like hidden representations \cite{tinybert}, attention dependencies \cite{minilm}, and relationships among representations \cite{distilling} are considered essential for effective knowledge transfer. In this paper, we introduce long context distillation (LCD) as the contextualized knowledge that primarily guides the student model. Specifically, the teacher model, FiD \cite{fid}, which processes longer contextual inputs, theoretically contains more information due to its richer context. This enables it to activate more specific internal knowledge, serving as a supervisory model. The teacher model aids the student model, T5 \cite{T5}, which is of the same size but uses shorter contextual inputs, in activating richer feature representations and knowledge.
The optimization objective for the student model at each mini-batch $z_r = (x_r, y_r)$ is:
\begin{equation}
\begin{split}
    \mathcal{L}_\text{s}(\theta_s, \theta_t, z_r) = \alpha \mathcal{L}_{\text{ce}}(y_r, S(x_r; \theta_s))\\ + (1 - \alpha) \mathcal{L}_{\text{ce}}(T(x_r; \theta_t), S(x_r; \theta_s))
\end{split}
\end{equation}
where we have a teacher model denoted as $T(\cdot; \theta_t)$ and a student model denoted as $S(\cdot; \theta_s)$. The corresponding model parameters are $\theta_t$ and $\theta_s$.

As illustrated on the right of Figure \ref{fig1}, we perform additional representation alignment to facilitate better knowledge transfer. In our distillation process, both the teacher and student models have \( L \) layers. The input text is processed through these layers, yielding corresponding output hidden states \(\{H_{l}^{t}\}_{l=0}^{L}\) and \(\{H_{l}^{s}\}_{l=0}^{L}\), along with attention matrices \(\{A_{l}^{t}\}_{l=1}^{L}\) and \(\{A_{l}^{s}\}_{l=1}^{L}\). 
% Assuming that the student's \(l\)-th layer aligns with the teacher's \(l\)-th layer, the student's outputs (\(H_{l}^{s}\) and \(A_{l}^{s}\)) should closely match those of the teacher (\(H_{l}^{t}\) and \(A_{l}^{t}\)). 
For aligning hidden states, we calculate the proximity between the teacher's and student's hidden states using cosine distance (COS) \cite{distilling}.
% As shown in the right of Figure \ref{fig1}, we perform an additional representation alignment for better knowledge transfer. In our distillation, both teacher model and student model have the $L$ layers, we feed the text into them and can obtain the corresponding output hidden states $\{H_{l}^{t}\}_{l=0}^{L}$, $\{H_{l}^{s}\}_{l=0}^{L}$, and attention matrices $\{A_{l}^{t}\}_{l=1}^{L}$, $\{A_{l}^{s}\}_{l=1}^{L}$. We suppose the student’s $l$-th layer is aligned with the teacher’s $l$-th layer, then the outputs of the student (i.e., $H_{l}^{s}$ and $A_{l}^{s}$) should be close to the teacher’s (i.e., $H_{l}^{t}$ and $A_{l}^{t}$). For aligning hidden states, following \cite{distilling}, we use cosine distance COS to calculate the proximity between the hidden states of the teacher and the student:
\begin{equation}
    \mathcal{L}_{\text{hid}} = -~\text{COS}(H_{l}^{s}, H_{l}^{t})
\end{equation}
While for aligning attention dependencies, we follow \cite{tinybert} to optimize the mean square error (MSE) between the attention matrices of the teacher and the student:
\begin{equation}
    \mathcal{L}_{\text{attn}} = -~\text{MSE}(A_{l}^{s}, A_{l}^{t})
\end{equation}
The overall objective for knowledge transfer is:
\begin{equation}
    \mathcal{L}_{\text{align}}(H_{l}^{s}, H_{l}^{t}, A_{l}^{s}, A_{l}^{t}) = \mathcal{L}_{\text{attn}} + \mathcal{L}_{\text{hid}}
\end{equation}
The overall objective for training AAG is the weighted sum of the two objectives:
\begin{equation}
    \mathcal{L} = \mathcal{L}_{\text{s}} + \lambda\mathcal{L}_{\text{align}}
\end{equation}

\begin{table*}[t]
\centering
% \vspace{-0.5em}
\renewcommand\arraystretch{1.05}
\resizebox{\linewidth}{!}{
\begin{tabular}{lccccccc}
\hline
      \multirow{2}{*}{\bfseries Models} & \multirow{2}{*}{\bfseries \# Docs} & \multicolumn{2}{c}{\textbf{NQ}} & \multicolumn{2}{c}{\textbf{TriviaQA}} & \multicolumn{2}{c}{\textbf{WebQ}} \\
    \cmidrule(r){3-4}\cmidrule(r){5-6}\cmidrule(r){7-8}
      &  & Large (800M) & XL (3B) & Large (800M) & XL (3B) & Large (800M) & XL (3B)\\
    \hline
    \rowcolor{gray!20} \multicolumn{8}{l}{\textit{\# Closed-book Setting}}\\
    T5 \cite{T5}  & 0 & 28.5$^*$ & 28.30 & 28.7$^*$ & 33.92 & 30.6$^*$ & 34.43 \\
    LoRA \cite{LoRA} & 0 & 17.70 & 23.15 & 23.87 & 32.16 & 29.13 & 35.24 \\
    AAG (Ours)  & 0 & \textbf{29.32} & \textbf{29.59} & \textbf{30.11} & \textbf{35.71} & \textbf{32.68} & \textbf{37.40} \\
    \hline
    \hline
    \rowcolor{gray!20} \multicolumn{8}{l}{\textit{\# Retrieval Augmented Setting (compared with RAG)}}\\
    DPR$^*$ \cite{DPR} (110M)  & 100 & 41.5 & - & 56.8 & - & 41.1 & - \\
    RAG$^*$ \cite{RAG} & 10 & 44.5 & - & 56.1 & - & 45.2 & - \\
    FiD$^*$ \cite{fid}  & 10 & 46.7 & 50.1 & 61.9 & 66.3 & 48.1 & 50.8  \\
    FiD \cite{fid} & 100 & 51.4$^*$ & \textbf{55.2}$^\ddagger$ & 67.6$^*$ & \textbf{72.9}$^\ddagger$ & 50.5 & \textbf{52.9}$^\ddagger$ \\
    EAR \cite{ear} & 10	& 39.6 & 42.3$^*$	& 60.0 & 64.6$^*$ & -	& - \\
    RFiD \cite{rfid} & 10	& 48.3 & 50.5	& 63.4 & 67.8 & -	& - \\
    FILCO$^*$ \cite{filco} 	& 1	 & - & 44.7	 & - & 59.0	& - & - \\
    AAG (Ours) & 10 & 49.9 & 50.9$^\ddagger$ & 69.7  & 70.3$^\ddagger$ & 51.5 & 52.8$^\ddagger$ \\
     AAG (Ours) & 30 & \textbf{53.1} & - & \textbf{70.5} & - & \textbf{52.0} & -\\
    \hline
    \rowcolor{gray!20} \multicolumn{8}{l}{\textit{\# Generation Augmented Setting (compared with GAG)}}\\
    GENREAD (sampling)$^*$ \cite{generate} & 10$^\dagger$ & 40.3 & 42.6 & 67.8 & 69.6 & 51.5 & 52.6 \\
    GENREAD (clustering)$^*$ \cite{generate} 	& 10$^\dagger$	& 43.5 & 45.6	& 70.2 & 71.6	& 53.5 	& 54.4 \\
    AAG (Ours)  & 10$^\dagger$ & \textbf{48.8} & \textbf{49.2}$^\ddagger$ & \textbf{70.9} & \textbf{72.2}$^\ddagger$ & \textbf{54.5} & \textbf{55.6}$^\ddagger$ \\
    \hline
    \rowcolor{gray!20} \multicolumn{8}{l}{\textit{\# Awakening Augmented Setting}}\\
    LoRA \cite{LoRA} & 1$^\dagger$ & 40.1 & 44.2 & 62.8 & 66.9 & 43.7 & 48.2 \\
    AAG (Ours)  & 1$^\dagger$ & \textbf{42.3}  & \textbf{46.5} & \textbf{65.5} & \textbf{68.4} & \textbf{45.3} & \textbf{50.5} \\
    \hline
\end{tabular}
}
\caption{QA performances of different methods with different settings. The first part (closed-book setting) indicates that only utilize questions; The latter three parts utilize explicit documents. The best results are in bold, while the second-best are underlined. * means that those results are from existing papers, $^\dagger$ denotes that the documents were generated ($^\ddagger$ indicates that the number of documents is reduced due to insufficient memory for distillation).}
% \vspace{-2.2em}
\label{tab1}
\end{table*}

\section{Experiment}
In this section, we conduct experiments to demonstrate the effectiveness and efficiency of AAG on QA. The experiment mainly answers four research questions (RQs):

\noindent RQ1: Can AAG achieve knowledge augmentation for QA over LLMs? (\cref{main})

\noindent RQ2: Does AAG have a good out-of-distribution generalization ability? (\cref{ood})

\noindent RQ3: Does AAG have advantages in effectiveness and efficiency compared to RAG and GAG? (\cref{cost})

\noindent RQ4: What is the role of explicit and implicit awakening modules in AAG? (\cref{ab})

\subsection{Datasets}
\label{sec:dataset}

We evaluate the proposed approach on three public question answering datasets: NaturalQuestions (\textbf{NQ}) \cite{nq}, WebQuestions (\textbf{WQ}) \cite{wq} and TriviaQA (\textbf{TQA}) \cite{tqa}. To evaluate the model performance, we use the exact match (EM) score for evaluating predicted answers \cite{squad}. We provide dataset details in the \ref{sec:appdataset}.

\subsection{Baselines}

Both the moderately sized language model (\textless 1B) and the large language model ($\geq$ 3B) are under consideration. T5 \cite{T5} is selected as the backbone for our moderately sized language models. We evaluate our proposed AAG against several knowledge-enhanced approaches, which include RAG models such as DPR \cite{DPR}, RAG \cite{RAG}, EAR \cite{ear}, RFiD \cite{rfid}, FILCO \cite{filco} and FiD \cite{fid}, as well as the GAG model GENREAD \cite{generate}, and parameters efficient fine-tuning method LoRA \cite{LoRA}.
% \ref{sec:base} provides further information about baselines.

To demonstrate the plug-and-play capability of AAG on the zero-shot settings of LLMs ($\geq$ 3B), we use Llama2-7B and -13B \cite{llama} as the basic model. We evaluate with 6 diverse settings: without retrieval, with retrieval, with LoRA, RECITE \cite{recitationaugmented}, HICL \cite{hintenhanced} and using the proposed AAG. 

\subsection{Implementations}

In the pretraining stage, the \textbf{context generator} initialized with T5-large utilizes the generated question-compressed pairs. During the second stage, the teacher model employs a FiD reader with different sizes (FiD-l and FiD-xl) that are fine-tuned on the training split of target datasets. The student model freezes the backbone and updates solely the hypernetwork, FFN and norm layers. \ref{sec:ex} contains more implementation and baseline details.
% , the feedforward neural network (FFN), and the normalization layers \cite{longLoRA}. The Appendix \ref{sec:ex} contains more implementation details and experimental findings.

\subsection{Main Results}
\label{main}

\subsubsection{Supervised Setting}
% Table~\ref{tab1} shows the performance results, full results including T5-Base are in the \ref{sec:Supervised}. When juxtaposed with closed-book models, RAG, and GAG methods, our proposed AAG framework AAG method exhibits state-of-the-art (SOTA) performance with the equivalent magnitude of document numbers. 

% In the closed-book setting (in the upper part of the table), our method outperforms the baseline by an average of \textbf{+2\% EM} score, indicating its excellence in utilizing internal knowledge with imagination. It’s especially noteworthy that as the model size expands, the performance advantages of the imagination become ever more evident.
Table~\ref{tab1} presents the performance results, with full results including T5-Base detailed in \ref{sec:Supervised}. Compared to closed-book models, as well as RAG and GAG methods, our proposed AAG method, achieves state-of-the-art (SOTA) performance using an equivalent number of documents.

In the closed-book setting (upper part of the table), our method surpasses the baseline by an average of +2\% EM score, demonstrating its superior ability to leverage internal knowledge through awakening. Notably, as the model size increases, the performance gains from the awakening approach become even more pronounced.

The following sections present the experimental results in the open domain setting\footnote{Due to memory constraints, AAG under the RAG setting using 30 documents.}. \textbf{Notably, proposed AAG using just one short dummy document, matches or exceeds the performance of RAG and GAG methods, which process 10 documents}. These results demonstrate that AAG effectively balances efficiency and overhead by leveraging imagined compressed text. 

\textbf{AAG outperforms baselines when documents-matched.} When AAG utilizes 10 retrieved documents under RAG setting, it surpasses RFiD performance by 1.6\% in NQ, 4.4\% in TQA, and 2.7\% in WQ. When AAG utilizes 10 generated documents under the GAG setting, it surpasses strong baseline GENREAD (clustering) performance by 4.5\% in NQ, 0.7\% in TQA, and 1.1\% in WQ.

\begin{figure}[t]
\centerline{\includegraphics[width=0.5\textwidth]{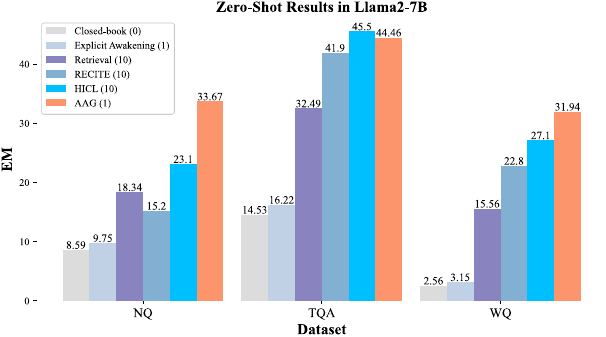}}
% \vspace{-0.45cm}
\caption{Zero-Shot results (EM, \%)  of Llama2-7B on three open-domain QA datasets. The number in parentheses indicates the number of documents used. More zero-shot setting results can be seen in \ref{sec:zero_sub}.}
\label{fig3}
% \vspace{-0.65cm}
\end{figure}

\subsubsection{Zero-shot Setting}
Figure \ref{fig3} illustrates the zero-shot results for LLMs implementing AAG with a frozen Llama2-7B and -13B. This research seeks to explore the possibility of enhancing LLMs via AAG. Due to the high computational demands of training, we only fine-tuned the hypernetwork on a mixed dataset without LCD in this experiment and evaluated performance in a zero-shot setting. Detailed prompt information can be found in the \ref{sec:prompts}. 

We discerned that Llama2's performance can be enhanced by imagining knowledge autonomously. While leveraging explicit imagined context could amplify the average EM +1\%, this is not as significant as the improvement achieved by retrieving 10 documents, indicating the limitations of relying solely on prompt cues for triggering corresponding knowledge. AAG can enhance knowledge via two main awakening processes, escalating EM by +15.33\% for NQ, +11.97\% for TQA, and +16.38\% for WQ. Compared to two other advanced RAG methods, AAG using a single document performs only 1 EM lower than the HICL method \cite{hintenhanced} on the TQA but achieves +10\% EM on the NQ and +5\% EM on the WQ. With AAG, Llama2-7B demonstrated an average improvement of +14\% across the three datasets. This trend is also observed in Llama2-13B's results (Figure \ref{zero13}). This implies that even in zero-shot settings, our method can still offer substantial benefits to LLMs.
% We discerned that Llama2's performance can be enhanced by autonomously imagining knowledge. While leveraging explicit imagined context could amplify the average EM by +1\%, this is not as significant as the improvement achieved by retrieving 10 documents, indicating the limitations of relying solely on prompt cues for triggering knowledge. AAG can enhance knowledge via two main imagination processes, escalating EM by +15.33\% for NQ, +11.97\% for TQA, and +16.38\% for WQ. Compared to two other advanced RAG methods, AAG, using a single document, performs only 1 EM lower than the HICL method \cite{hintenhanced} on the TQA dataset, but achieves over 10 EM higher on the NQ dataset and over 5 EM higher on the WQ dataset. With AAG, Llama2-7B demonstrated an average improvement of +14\% across the three datasets. This trend is also observed in Llama2-13B's results\footnote{More results can be seen in \ref{sec:zero_sub}.}. This implies that even in zero-shot settings, our method can still offer substantial benefits to LLMs. 

\begin{table}[t]
\centering
% \vspace{-0.5em}
\renewcommand\arraystretch{1.05}
\resizebox{\linewidth}{!}{
\begin{tabular}{lccccccc}
\hline
    \multirow{2}{*}{\bfseries Models} & \multirow{2}{*}{\bfseries \# Docs} & \multicolumn{3}{c}{\bfseries Base (220M)} & \multicolumn{3}{c}{\bfseries Large (800M)}\\
    \cmidrule(r){3-5}\cmidrule(r){6-8}
      &  & {\bfseries \underline{NQ}} & {\bfseries TQA} & {\bfseries WQ} & {\bfseries \underline{NQ}} & {\bfseries TQA} & {\bfseries WQ} \\
\hline
    T5 & 0 & 22.16 & 3.18 & 4.12 & 28.5* & 3.18 & 4.12 \\
    AAG & 0 & 23.89 & 6.21 & 10.94 & 29.32 & 10.17 & 14.06 \\ \hline
    FiD & 10 & 46.81 & 53.93 & 24.02 & 46.7* & 57.93 & 25.12\\
    AAG & 10 & \textbf{47.01} & \textbf{55.74} & \textbf{24.13} & \textbf{49.92} & \textbf{60.03} & \textbf{25.79} \\ \hline
    LoRA & 1$^\dagger$ & 37.17 & 45.20 & 15.62 & 37.61 & 48.50  & 20.71 \\
    AAG & 1$^\dagger$ & 40.14 & 46.61 & 18.92 & 42.32 & 54.80 & 22.05\\ 
\hline
\end{tabular}
}
\caption{\textbf{IID and OOD results.} The performance on three open-domain datasets for the model trained on NQ is reported, with underlined values indicating IID performance. Full OOD results and details of the three datasets are provided in the \ref{sec:ood}.}
% \vspace{-2.2em}
\label{tab2}
\end{table}

% Moreover, the proposed method does not benefit from parameter-efficient fine-tuning (LoRA), but rather from the superiority of each module. 
\subsection{Out-Of-Distribution (OOD) Performance}
\label{ood}

To further demonstrate the generalizations of the AAG method and the importance of hypernetwork, we also evaluate its performance in OOD generalizations. Table \ref{tab2} shows the IID and OOD performance of FiD, and AAG methods with different document settings when training on NQ (From NQ generalization to the other two datasets). 

It is patently clear that an increment in document provision leads to better OOD performance, likely due to the presence of answer-oriented content within these documents. Remarkably, AAG can come within a relatively narrow 5\% gap of FiD, even when utilizing a single imagined document as opposed to 10 retrieved documents. 

Simultaneously, AAG generally showcases superior performance in OOD when provided with 10 retrieved documents. This superiority can be traced back to the pivotal role played by hypernetwork in generating LoRA adapters' weights based on questions. This equips models with the capability to invoke and access internal knowledge based on context-specific discourse rather than confining to resolving distinct questions.

% \begin{table}[t]
% \centering
% % \vspace{-0.5em}
% \renewcommand\arraystretch{1.05}
% \resizebox{\linewidth}{!}{
% \begin{tabular}{lC{4.5em}C{3.5em}C{5em}C{5em}C{5em}C{5em}}
% \hline
%     \textbf{Models} & \textbf{Training Params} & \textbf{\# Documents} & \textbf{\# Avg Tokens} & \textbf{Inference Time} & \textbf{GPU Memory} & \textbf{Training Time} \\ \hline
%     T5 & 220M & \textbf{0} & \textbf{19.8} & \textbf{79.8s} & 2828M & \textbf{0.9h} \\
%     AAG & \textbf{139.3M} & \textbf{0} & \textbf{19.8} & 82.3s & \textbf{2710M} & 1.2h \\
%     % \hline
%     % LoRA & 141.5M & 1 & 482 & 702.9s & 2716M & 1.6h \\
%     AAG & \textbf{139.3M} & 1 & 522.1 & 214.6s & 2882M  & 1.7h\\
%     \hline
%     FiD & 220M & 10 & 1748.3 & 683.3s & 4358M  & 2.3h\\
%     GENREAD & 220M & 10 & 1912.5 & 704.8s & 4412M  & -\\
%     FiD & 220M & 100 & 16625.7 & 1293.2s & 19048M  & 5.8h\\ \hline
% \end{tabular}
% }
% \caption{Training and inference cost on the NQ. The backbone model is T5-Base.}
% % \vspace{-2.2em}
% \label{tab3}
% \end{table}

\begin{table}[t]
\centering
% \vspace{-0.5em}
\renewcommand\arraystretch{1.1}
\resizebox{\linewidth}{!}{
\begin{tabular}{lC{4.5em}C{3em}C{5em}C{5em}C{5em}}
\hline
    \textbf{Models} & \textbf{Training Params} & \textbf{\# Docs} & \textbf{\# Avg Tokens} & \textbf{Inference Time} & \textbf{Training Time} \\ \hline
    T5 & 220M & \textbf{0} & \textbf{19.8} & \textbf{79.8s} & \textbf{0.9h} \\
    AAG & \textbf{139.3M} & \textbf{0} & \textbf{19.8} & 82.3s  & 1.2h \\
    % \hline
    % LoRA & 141.5M & 1 & 482 & 702.9s & 2716M & 1.6h \\
    AAG & \textbf{139.3M} & 1 & 522.1 & 214.6s  & 1.7h\\
    \hline
    FiD & 220M & 10 & 1748.3 & 683.3s   & 2.3h\\
    GEN. & 220M & 10 & 1912.5 & 704.8s   & -\\
    FiD & 220M & 100 & 16625.7 & 1293.2s   & 5.8h\\ \hline
\end{tabular}
}
\caption{Training and inference cost on the NQ.}
% \vspace{-2.2em}
\label{tab3}
\end{table}

\subsection{Training Cost and Inference Speed-up}
\label{cost}
We proceeded to measure the inference speed documented in GPU time and training time for 5000 steps on the NQ dataset using T5-Base. The experiments were conducted on a single RTX 3090 GPU, maintaining a standard batch size of 8 during training and 1 during inference. A detailed inference case is shown in the Appendix \ref{case}.

% As evident from Table \ref{tab3}, the proposed method's advantage lies in its diminished requirement for parameter updates, attributed to the shared hypernetwork's utilization that generates adapters, thus negating the need for individual LoRA adapters' setup. Despite lacking a training advantage due to distillation constraints, AAG achieves efficient reasoning through an extremely lightweight design, saving more than half the training time compared to methods using a large number of documents ($\times0.3$). Compared to the other two methods, the processed tokens are significantly decreased, while either outperforming them or showing negligible differences in performance. This represents an optimal trade-off between efficiency and computational demand. Moreover, unlike GAG, our approach incurs no financial costs associated with API calls, and the reduced model size facilitates faster generation.
As evident from Table \ref{tab3}, the proposed method's advantage lies in its diminished requirement for parameter updates, which can be attributed to the shared hypernetwork's utilization that generates LoRA adapters, thereby negating the necessity of individual LoRA adapters' setup. Despite the lack of a training advantage due to distillation constraints, AAG achieves efficient reasoning through an extremely lightweight design, saving more than half the training time compared to methods using a large number of documents ($0.3\times$). Compared to the other two methods, the processed tokens are significantly decreased, while either outperforming them or showing negligible differences in performance. This represents an optimal trade-off between efficiency and computational demand. Moreover, unlike GAG, our approach incurs no financial costs associated with API calls, and the reduced model size facilitates faster generation.

\subsection{Ablation Study}
\label{ab}

This study introduces two key awakening processes to stimulate LLMs' internal knowledge: explicit awakening (EA) and implicit awakening (IA). We particularly examined the influence of different awakening types on performance.

Table \ref{ta:ab} demonstrates that both EA and IA are important for AAG. Omitting either one results in a considerable reduction in performance, with a drop exceeding 30\% observed when EA is neglected. This is harmonious with the initial observation that performance improvement becomes more noticeable when relevant documents are available, thus underscoring EA's superiority. 

The outcomes of Long Context Distillation (LCD) including $\mathcal{L}_{s}$ and $\mathcal{L}_{align}$ also make marginal contributions to the overall results. This validates the previous assertion that a more extensive context tends to optimize performance, although with limited gains. The impact of EA on the application of hypernetworks is minimal (\textless 3\%), indicating that hypernetworks in IA primarily serve to awaken parameter knowledge rather than to utilize the generated context. The experiments and analysis above demonstrate the importance of each component and the effectiveness of our AAG method.

\begin{table}[t]
\centering
% \vspace{-0.5em}
\renewcommand\arraystretch{1.05}
\resizebox{\linewidth}{!}{
\begin{tabular}{lcccc}
\hline
    \bfseries Methods & \bfseries \# Docs ($\downarrow$) & NQ ($\uparrow$) & TQA ($\uparrow$) & WQ ($\uparrow$) \\
\hline
    AAG & 1$^\dagger$ & 40.14 & 60.75 & 41.73 \\ 
    w/o EA & 0 & 23.89 ($\downarrow 40\%$) & 22.69 ($\downarrow 63\%$) & 30.31 ($\downarrow 27\%$)\\
    ~~In. w/o EA & 1 & 38.85 ($\downarrow 3\%$)~~ & 59.62 ($\downarrow 2\%$)~~ & 40.65 ($\downarrow 3\%$)~~~\\
    w/o IA & 1 & 33.48 ($\downarrow 17\%$) & 51.19 ($\downarrow 16\%$) & 34.72 ($\downarrow 17\%$)\\
    w/o LCD & 1 & 33.96 ($\downarrow 15\%$) & 53.27 ($\downarrow 12\%$) & 29.39 ($\downarrow 29\%$)\\
    ~~w/o $\mathcal{L}_s$ & 1 &  34.24 ($\downarrow 14\%$) & 54.90 ($\downarrow 10\%$) & 31.67 ($\downarrow 24\%$)\\
    ~~w/o $\mathcal{L}_{align}$ & 1 &  37.41 ($\downarrow 7\%$)~~ & 56.38 ($\downarrow 7\%$)~~ & 39.26 ($\downarrow 6\%$)~~~\\
\hline
\end{tabular}
}
\caption{Ablation studies on three open-domain QA datasets. The backbone model is the T5-base. "In." means the input of the hypernetwork \cref{ssec:Hypernetwork}.}
% \vspace{-2.2em}
\label{ta:ab}
\end{table}

% \begin{figure}[t]
% \centerline{\includegraphics[width=0.5\textwidth]{ab.pdf}}
% % \vspace{-0.45cm}
% \caption{Ablation experiment results (\%) of T5-Base on three open-domain QA datasets.}
% \label{fig4}
% % \vspace{-0.65cm}
% \end{figure}

\section{Conclusion}

This study proposes a novel knowledge-augmented strategy for Large Language Models (LLMs), namely Awakening Augmented Generation (AAG) for open domain question answering. The AAG effectively harnesses the inherent knowledge of the LLMs through a dual-awakening approach to awaken a richer context. Explicit awakening with the context generator generates a short dummy document as symbolic context, while implicit awakening uses hypernetwork to convert the question and the document into adapters inserted into the LLMs as parameter context. Experimental results demonstrate a significant improvement in performance while remaining relatively lightweight. Although the main focus of this method is on one specific task, we believe these findings can offer a novel perspective on how to better harness the potential of LLMs. 
% In the future, we plan to apply AAG to more NLP tasks and explore multimodal knowledge-augmented generation.

\section*{Limitations}

While this study has demonstrated significant achievements in QA tasks, there are notable limitations:

\noindent \textbf{Tasks.}~~ The proposed methods in the study are specialized specifically for QA. It remains unknown how effective they would be in other types of knowledge-intensive tasks, such as fact-checking or dialogue systems. Further validation is needed to assess the generalizations and applicability of this approach.

\noindent \textbf{Multimodal.}~~ We have only considered imagined text and hidden representations. In future work, it is imperative to explore multimodal information including the impact of imagining images on performance.

\noindent \textbf{Method.}~~ Our method relies on the knowledge learned by LLMs in the pre-training phase, which may limit the model's ability to quickly adapt to new information. The dependency on internal knowledge activation in AAG may lead to a less transparent decision-making process in the model, making it challenging to explain the logic behind the generated answers. In the future, there is a need to continue exploring adaptive knowledge enhancement methods to optimize results further.

\noindent \textbf{Hypernetwork.}~~ For lightweight and efficient settings, our hypernetwork employs a two-layer MLP. However, some studies use larger models, such as GPT-2 or T5, as hypernetworks. Due to computational resource constraints, we did not explore or compare the effects of different hypernetwork models on the results. Nonetheless, our method primarily focuses on generating parameter-efficient modules to enhance knowledge activation and generalization.

\section*{Ethical Considerations}
In this paper, we proposed a novel knowledge enhancement method aimed at leveraging the knowledge of LLMs. However, LLMs may generate inappropriate or discriminatory knowledge. Our approach does not introduce ethical concerns. The datasets we used are public, and there are no privacy issues.
% \section*{Acknowledgements}

\section*{Acknowledgements}

This work was supported by National Key R\&D Program of China (No. 2022YFF0711900) and the National Natural Science Foundation of China (No.62376270, No.62276264). This work was supported by the Youth Innovation Promotion Association CAS.
\bibliography{custom}

\appendix

\begin{table*}[h]
    \centering
    \resizebox{\linewidth}{!}{
    \begin{tabular}{lcccc}
        \toprule
        & Document Relevance & Context Length & Inference Time & Inference Dependence \\
        \midrule
        RAG & Medium & Too Long & Very High & Retriever \\
        GAG & High & Long & High & Larger Model (InstructGPT) \\
        AAG & High & Short & Low & None \\
        \bottomrule
    \end{tabular}
    }
    \caption{Comparison of Different Paradigms}
    \label{tab:comparison_methods}
\end{table*}

\section{Method}
\subsection{Comparison of Three Paradigms}
\label{method_com}
Compared to RAG and GAG, our method has certain limitations, such as requiring a more complex training process and the necessity of training a model. Similar to the GAG method, which uses a master’s degree in law as a knowledge base, our method also struggles to generate content when encountering new and unknown world knowledge, which presents a challenge that needs to be addressed. Additionally, the knowledge base might be affected by knowledge gaps in low-resource settings where there is a lack of a comprehensive knowledge base.

Next, we compare AAG, RAG, and GAG across four criteria for a more intuitive understanding. From the table \ref{tab:comparison_methods}, it can be observed that the document relevance obtained by AAG and GAG is higher, while RAG heavily relies on the retriever and external knowledge base. In terms of document length usage, AAG only needs to use a virtual document, greatly reducing the number of tokens. Therefore, AAG is superior to the other two methods in terms of reasoning time.

\subsection{Context Generator}
\label{role}
There are two main goals in the pre-training of the model in the first stage of AAG (context generator): first, to improve its ability as a document generator by learning to generate rich and concise documents; second, to introduce some external knowledge that the model initially does not possess. It’s worth noting that the second goal is crucial, as the model may encounter knowledge it has not yet learned. Thus, AAG does not rely on external large models or retrievers for external reasoning and can complete reasoning independently.

\subsection{Hypernetwork}
\label{reason}

Hypernetworks have gained significant attention in recent years due to their potential to enhance various aspects of neural network performance. In this section, we analyze the reasons for employing hypernetworks in detail:

Hypernetworks \cite{hypernetworks} offer a solution that reduces the dependency on gradient descent for specific domains. Methods such as Hypertuning \cite{hypertuning} and HINT \cite{hint} use hypernetworks to transform inputs into parameter-efficient modules, thereby reducing computation and enhancing model generalization.

Hypernetworks, which are neural networks designed to generate the weights of other networks, allow for dynamic adjustment of model parameters. This adaptability enables the model to better suit different tasks and datasets, thereby improving overall performance. By utilizing hypernetworks, the number of models that need to be trained individually can be significantly reduced. Traditional methods require separate models for each task, whereas hypernetworks can generate weights for multiple tasks. This capability enhances training efficiency. In our task, we use hypernetworks to generate adapters for the question and input, which are then inserted into the model. This helps the model incorporate the knowledge targeted by the question, corresponding to implicit awakening. Compared to traditional efficient fine-tuning, this process is more aligned with the goal of awakening.

Hypernetworks can capture the commonalities and differences between various tasks by learning to generate weights. This ability to generalize across tasks improves the model's performance on unseen data, making it more robust in diverse scenarios. In multi-task learning or meta-learning scenarios, hypernetworks can considerably reduce the need for storing multiple independent models. A hypernetwork only needs to store a single generating network and some shared parameters, thus significantly decreasing the storage space required. Hypernetworks can quickly generate new weights to adapt to new tasks as they arise. This rapid adaptation capability is particularly useful in applications that require frequent updates or expansions. In our experiments \ref{ood}, we also found that using a hypernetwork can significantly enhance the generalization ability for tasks. This is because it not only retains knowledge within the domain-specific modules but also learns to generate question-targeted knowledge to be inserted into the model.

\section{Experimantal Settings}
\label{sec:appendix}

\subsection{Background}
\label{sec:back}

Our task formulation follows retrieval augmented models for QA \cite{REALM, end}. Let $\mathcal{V}^*$ denote the infinite set, encompassing all potential strings over the tokens in vocabulary $\mathcal{V}$, and this includes the empty string. An instance within a QA dataset is defined as a triplet $(\boldsymbol{q}, \boldsymbol{a}, \boldsymbol{c})$ comprising question $\boldsymbol{q}$, answer $\boldsymbol{a}$, and context $c$, where $\boldsymbol{q}, \boldsymbol{a}, \boldsymbol{c} \in \mathcal{V}^*$. Conventionally, the context $c$ is drawn from the knowledge corpus $\mathcal{Z}$, like Wikipedia, whereby $\mathcal{Z} \subset \mathcal{V}^*$.

The goal of QA is to learn a distribution function, represented as \( p(\boldsymbol{a}|\boldsymbol{q}) \), wherein the models decode a string $\boldsymbol{a}$ that serves as an abstractive answer to a given query $\boldsymbol{q}$. In a closed-book setting, LLMs directly encode the given question and predict the answer \cite{much}. Specifically, considering the context $c$ as the empty string, the reliance is solely on the model parameters, i.e., $\hat{\boldsymbol{a}} = \arg\max_{\boldsymbol{a} \in \mathcal{V}^*} p(\boldsymbol{a}|\boldsymbol{q},\theta)$, where $\theta$ represents the LLMs' parameters. However, employing a direct approach of requesting models to output answers frequently results in subpar performance, primarily attributable to omitting a substantial amount of world knowledge during the process. Therefore, a popular approach is open domain setting, which marginalizes \( p(\boldsymbol{a}|\boldsymbol{q}, \boldsymbol{c}) \) over contexts $c$ in the knowledge corpus \cite{RAG, end} or generated from models \cite{generate}. Given the computational infeasibility of calculating probabilities for all contexts, $p(\boldsymbol{a}|\boldsymbol{q}, \boldsymbol{c})$ is approximated to the sum of probabilities for top $k$ contexts, i.e., $p(\boldsymbol{a}|\boldsymbol{q}, \boldsymbol{c}) = \sum\limits^{\boldsymbol{c}_i \in \boldsymbol{c}}\limits_{c \in \text{Topk}(\boldsymbol{q})} p(\boldsymbol{a}|\boldsymbol{q},\boldsymbol{c}_i)p(\boldsymbol{c}_i|\boldsymbol{q})$, where $\text{Topk}(\boldsymbol{q})$ denotes the set of resulting top $k$ passages after the retrieval or generated with a query $\boldsymbol{q}$.

\subsection{Prompts for Explicit Imagine with LLMs}
\label{sec:prompts}

The prompt for explicit awakening of the context generator to imagine a short dummy useful document is:

\textit{Imagine contexts based on the question: \textbackslash n {{input}} \textbackslash n Contexts: \textbackslash n}

Table \ref{prompts} shows the full prompts for zero-shot results on LLM that we use for open domain QA: NQ, TQA, WQ.

\begin{table}[htbp]
\centering
% \vspace{-0.5em}
\renewcommand\arraystretch{1.05}
\resizebox{\linewidth}{!}{
\begin{tabular}{lcccc}
\hline
    \textbf{Models} & \textbf{\makecell[c]{Docu-\\ments}} & \textbf{Steps} & \textbf{Lr} & \textbf{\makecell[c]{Batch\\Size}} \\ \hline
    T5 & 0 & 40000 & 1e-4 & 8 \\
    LoRA-Base & 0 & 40000 & 5e-4 & 8 \\
    AAG & 0 & 50000 & 1e-3 & 8 \\
    LoRA-l & 0 & 40000 & 1e-4 & 4 \\
    AAG-l & 0 & 50000 & 5e-4 & 4 \\
    FiD-3b & 0 & 40000 & 1e-4 & 2 \\
    LoRA-3b & 0 & 40000 & 1e-4 & 4 \\
    AAG & 0 & 50000 & 1e-4 & 1 \\ \hline
    LoRA-Base & 0$^\dagger$ & 40000 & 5e-4 & 8 \\
    AAG & 0$^\dagger$ & 50000 & 1e-3 & 8 \\
    LoRA-l & 0$^\dagger$ & 40000 & 1e-4 & 4 \\
    AAG-l & 0$^\dagger$ & 50000 & 5e-4 & 4 \\
    LoRA-3b & 0$^\dagger$ & 40000 & 1e-4 & 2 \\
    AAG-3b & 0$^\dagger$ & 50000 & 1e-4 & 1 \\ \hline
    AAG & 10 & 50000 & 5e-4 & 1 \\
    AAG-l & 10 & 50000 & 5e-4 & 1 \\
    FiD-3b & 10 & 40000 & 1e-4 & 1 \\
    AAG-3b & 10 & 50000 & 1e-4 & 1 \\ \hline
\end{tabular}
}
\caption{Hyperparameter Settings.}
% \vspace{-2.2em}
\label{set}
\end{table}

\begin{table*}[htbp]
\centering
% \vspace{-0.5em}
\renewcommand\arraystretch{1.05}
\resizebox{\linewidth}{!}{
\begin{tabular}{lC{3.5em}C{2.5em}C{2.5em}C{2.5em}C{2.5em}C{2.5em}C{2.5em}C{2.5em}C{2.5em}C{2.5em}}
\hline
    \multirow{2}{*}{\bfseries Models} & \multirow{2}{*}{\bfseries \# Docs} & \multicolumn{3}{c}{\bfseries NQ} & \multicolumn{3}{c}{\bfseries TQA} & \multicolumn{3}{c}{\bfseries WQ} \\ 
  &  & \underline{NQ} & TQA & WQ & NQ & \underline{TQA} & WQ & NQ & TQA & \underline{WQ} \\
\hline
    T5 & 0 & 22.16 & 3.18 & 4.12 & 2.65 & 21.8 & 3.15 & 0.88 & 2.95 & 28.3\\
    LoRA-Base & 0 & 16.17 & 4.71 & 6.89 & 3.15 & 21.16 & 0.00 & 1.33 & 3.04 & 26.38 \\
    AAG & 0 & 23.89 & 6.21 & 10.94 & 5.31 & 22.69 & 6.30 & 3.23 & 5.10 & 30.31 \\
    LoRA-Base & 1$^\dagger$ & 37.17 & 45.20 & 15.62 & 19.57 & 55.37 & 12.50 & 14.15 & 30.89 & 28.88 \\
    AAG & 1$^\dagger$ & 40.14 & 46.61 & 18.92 & 24.78 & 60.75 & 12.82 & 17.70 & 35.24 & 41.06 \\
    FiD & 10 & 46.81 & 53.93 & 24.02 & 28.57 & 63.32 & 17.83 & 18.81 & 41.88 & 41.78 \\
    AAG & 10 & \textbf{47.01} & \textbf{55.74} & \textbf{24.13} & \textbf{31.77} & \textbf{64.95} & \textbf{19.52} & \textbf{24.43} & \textbf{48.10} & \textbf{46.36} \\
\hline
    T5-l & 0 & 28.5* & 3.18 & 4.12 & 2.65 & 28.7* & 3.15 & 0.88 & 2.95 & 30.6*\\
    LoRA-l & 0 & 17.70 & 7.49 & 8.66 & 3.54 & 23.87 & 4.72 & 0.00 & 5.65 & 29.13 \\
    AAG-l & 0 & 29.32 & 10.17 & 14.06 & 7.02 & 30.11 & 7.81 & 2.65 & 7.06 & 32.68 \\
    LoRA-l & 1$^\dagger$ & 37.61 & 48.50 & 20.71 & 20.54 & 62.71 & 14.81 & 15.36 & 33.83 & 39.37 \\
    AAG-l & 1$^\dagger$ & 42.32 & 54.80 & 22.05 & 26.11 & 65.48 & 18.11 & 18.58 & 47.46 & 45.28 \\
    FiD-l & 10 & 46.7* & 57.93 & 25.12 & 34.29 & 61.9* & 19.64 & 27.65 & 53.87 & 48.1* \\
    AAG-l & 10 & \textbf{49.92} & \textbf{60.03} & \textbf{25.79} & \textbf{34.35} & \textbf{69.67} & \textbf{20.28} & \textbf{30.19} & \textbf{54.94} & \textbf{51.52} \\
\hline
\end{tabular}
}
\caption{\textbf{OOD results}. The primary row in the table header delineates the dataset trained, while the underscored secondary row demonstrates the in-distribution performance. AAG attains optimal performance both in-distribution and OOD under diverse document configurations.}
% \vspace{-2.2em}
\label{fullood}
\end{table*}

\subsection{Implementations}
\label{sec:ex}
In this section, we describe the implementation of our experiments in detail, including the baseline methods, backbone models, and hyperparameters. Our model is built based on the T5 \cite{T5}. Differing from fine-tuning all model parameters $\theta$ of the updated Pre-trained Language Model (LLM), LoRA \cite{LoRA} freezes all pre-trained Transformer parameters and optimizes only the parameters of each LoRA adapter. We employ LoRA to train a parameter-efficient fine-tuning baseline. Drawing from this, our approach updates only the parameters of the Hypernetwork to generate the weights for each LoRA adapter. This method is adopted based on LongLoRA's \cite{longLoRA} recommendations and experimental findings, demonstrating improved performance when the normalization and FFN layers components are updated. This is because: 1) dynamically generating LoRA weights enhances generalization and parameter sharing, and 2) LoRA performs comparably to fine-tuning but mitigates the risk of catastrophic forgetting.

For the baseline, most of the hyperparameters are the default parameters of FiD \cite{fid}. For LoRA \cite{LoRA}, add the LoRA module only to the $\mathcal{QV}$ of the attention layers and also release the normalization and FFN layers. 

We consider conducting experiments using three different sizes of T5, namely T5-base, T5-large, T5-3b, and Llama2-7B, Llama2-13B \cite{llama}. Due to memory constraints and online distillation limitations, A100 supports processing 20 documents for T5-3b, while Llama2 does not support distillation. All experiments with T5-3b are conducted on 2 A100 GPUs, T5-large on 2 A6000 GPUs, and T5-Base on 2 RTX 3090 GPUs. However, experiments with Llama2-7b and 13b, except for AAG on 2 A100 GPUs, are tested on 8 RTX 3090 GPUs.

\subsubsection{Hyperparameters}

The detailed hyperparameter setting is as shown in Table \ref{set}. For the LoRA modules, we set the $\alpha$ 32 and the \textit{lora rank} 32.

\subsubsection{Baselines}
\label{sec:base}
\noindent \textbf{DPR}~~ \cite{DPR} generates by searching for the most relevant documents through dense vector space representation.

\noindent \textbf{FiD}~~ \cite{fid} retrieve relevant documents and send them separately to the Encoder, then fuse the information in the Decoder.

\noindent \textbf{RFiD}~~ \cite{rfid} uses the encoder of FiD to distinguish between causal and incidental features, and guides the decoder to generate answers based on this distinction.

\noindent \textbf{EAR}~~ \cite{ear} significantly enhances the traditional sparse retrieval method BM25 by connecting query expansion models and retrievers.

\noindent \textbf{FILCO}~~ \cite{filco} identifies useful context based on lexical and information-theoretic methods.

\noindent \textbf{GENREAD}~~ \cite{generate} prompt LLMs like InstructGPT \cite{training} to generate a large number of relevant documents and let the reader process them.

\noindent \textbf{LoRA}~~ We use LoRA \cite{LoRA} to obtain an efficiently fine-tuned baseline and compare it with our method.

\subsubsection{Evaluation}
For QA datasets, we choose the exact match (EM) score \cite{squad} as the evaluation metric. An answer is deemed correct if it aligns with any of the responses in the list of acceptable answers after normalization. Normalization involves transforming the text into lowercase, omitting articles, punctuation, and eliminating redundant spaces.

\begin{table}[htbp]
    \centering
    \begin{tabular}{lccc}
        \toprule
        Dataset & Train & Dev & Test \\
        \midrule
        WebQ & 3,417 & 361 & 2,032 \\
        NQ   & 79,168 & 8,757 & 3,610 \\
        TQA  & 78,785 & 8,837 & 11,313 \\
        \bottomrule
    \end{tabular}
    \caption{Open-Domain QA dataset statistics}
    \label{tab:dataset_sizes}
\end{table}

\subsection{Downstream Evaluation Datasets}
\label{sec:appdataset}
We use the following three Open-Domain QA for the experiments (\cref{sec:dataset}).

\begin{itemize}
    \item NaturalQuestions \cite{nq} contains questions corresponding to Google search queries. The open-domain version of this dataset is obtained by discarding answers with more than 5 tokens, each accompanied by a Wikipedia article containing the answer.
    \item TriviaQA \cite{tqa} contains questions gathered from trivia and quiz-league websites. The unfiltered version of TriviaQA is used for open-domain question answering, each question is accompanied by pages from web and Wikipedia searches that may contain the answer.
    \item WebQuestions \cite{wq} contains questions from web queries matched to corresponding entries in FreeBase \cite{freebase}.
\end{itemize}

Table \ref{tab:dataset_sizes} presents detailed statistics of the dataset sizes, including the training, development, and test sets. We note that all our models are trained exclusively on the training data, and we did not include the development data in our training process. Therefore, the performance numbers reported in the paper for the dev and test data are independent of the training data.

\begin{figure}[t]
\centerline{\includegraphics[width=0.5\textwidth]{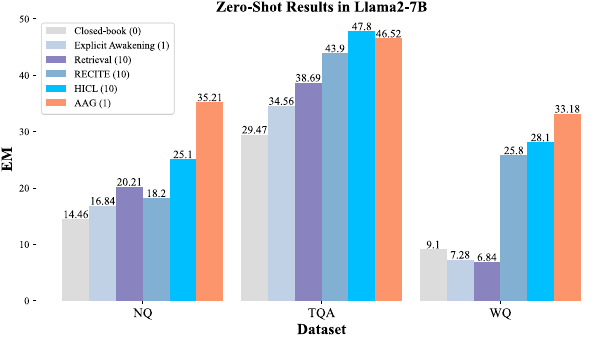}}
% \vspace{-0.45cm}
\caption{Zero-Shot results (EM, \%) of Llama2-13B on three open-domain QA datasets. The number in parentheses indicates the number of documents used.}
\label{zero13}
% \vspace{-0.65cm}
\end{figure}

\section{Full Experimental Results}

\subsection{Supervised Performance}
\label{sec:Supervised}

As shown in Table \ref{fullsup}, our initial observations indicate that regardless of the method implemented, supplying a certain quantity of related documents can expedite improvement and enhance performance in QA. FiD \cite{fid} model outclasses all baseline models in performance. Notably, utilizing FiD-xl with a mere 10 documents yields performance on par with that attained through the use of FiD-l with 100 documents. Larger models not only encapsulate more knowledge but also demonstrate a superior ability to activate and apply this knowledge efficiently.

Additionally, in comparison with LoRA \cite{LoRA} methods, AAG enhances EM scores by an average of +2.2\%. In the closed-book setting, the LoRA method manifests a substantial decrease in performance, likely attributable to the inadequacy of learning sufficient knowledge via questions for storage in the LoRA module. On the other hand, AAG harnesses both explicit and implicit awakenings to exploit knowledge for improved outcomes. These results indicate that the knowledge stored in the LLMs' parameters can still be further exploited.

\begin{figure}[t]
\centerline{\includegraphics[width=0.5\textwidth]{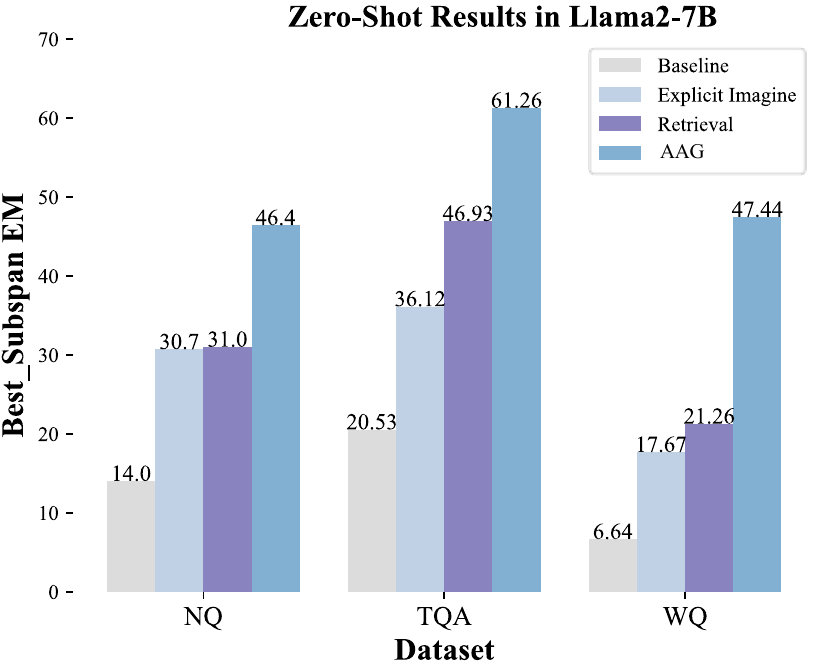}}
% \vspace{-0.45cm}
\caption{Zero-Shot results (Best\_Subspan EM, \%) of Llama2-7B on three open-domain QA datasets.}
\label{zerosub}
% \vspace{-0.65cm}
\end{figure}

\subsection{OOD Results}
\label{sec:ood}

Table \ref{fullood} shows the full OOD results in QA. It can be observed that our method has the best OOD generalization ability on all three benchmarks. Although LoRA performs well on the in-distribution part, its performance is generally poor on OOD, with some even showing negative performance. This highlights the importance of the domain adaptability of the implicit awakening Hypernetwork in our method, which generates LoRA adapter weights based on input.

\subsection{Zero-Shot Results}
\label{sec:zero_sub}
 LLMs have limited capacity to utilize extensive context effectively and are prone to generating illusions and redundant content. Best\_subspan EM assesses whether the answer is included in the output. Previous studies have corroborated that LLMs encapsulate a considerable volume of knowledge and exhibit robust performance in QA.
 
\begin{figure}[t]
\centerline{\includegraphics[width=0.5\textwidth]{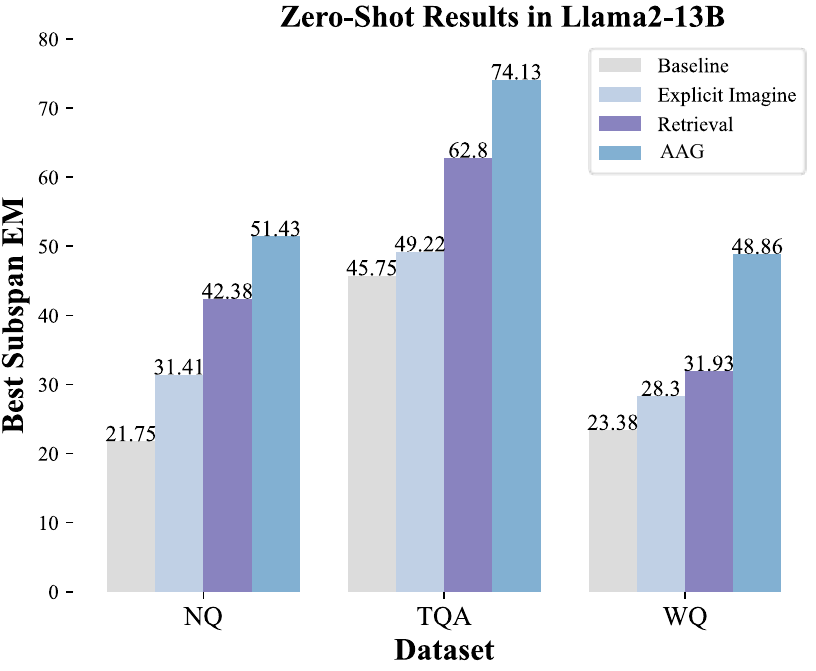}}
% \vspace{-0.45cm}
\caption{Zero-Shot results (Best\_Subspan EM, \%) of Llama2-13B on three open-domain QA datasets.}
\label{zerosub13}
% \vspace{-0.65cm}
\end{figure}

Here, we report the Best\_Subspan\_EM values of Llama2-7B and Llama2-13B on three QA datasets. From Figure \ref{zerosub} and Figure \ref{zerosub13}, it can be observed that Best\_Subspan\_EM significantly improves, but the EM values are relatively small. This indicates that LLMs may not effectively utilize retrieval documents and are prone to outputting a lot of irrelevant information. Therefore, there is an urgent need to explore efficient techniques that leverage external information and internal knowledge.

However, the model did exhibit a weak adherence to instructions, often failing to output the exact answer. Remarkably, Llama2-13B displayed a decline in EM with increased document length on the WQ dataset, whereas the Best\_Subspan\_EM value augmented. Contrarily, our method excelled in extracting key information by using text awakening during the compression phase.

\begin{table}[h!]
    \centering
    \resizebox{\linewidth}{!}{
    \begin{tabular}{lm{1.5cm}m{1.5cm}m{1.5cm}}
        \hline
        \textbf{Model} & \textbf{NQ} & \textbf{TriviaQA} & \textbf{WebQ} \\ \hline
        \multicolumn{4}{c}{\textit{\textbf{\# LLaMA-2-7B}}} \\ 
        Zero-shot & 8.6 & 14.5 & 2.6 \\ 
        DPR + ICL & 18.3 & 32.5 & 15.6 \\ 
        DPR + RECITE \cite{recitationaugmented} & 16.8 & 43.9 & 24.8 \\ 
        DPR + HICL \cite{hintenhanced} & 25.1 & 47.5 & 28.1 \\
        DPR + AAG (Ours) & 33.7 & 44.5 & 31.9 \\ 
        \hline
    \end{tabular}
    }
    \caption{Zero-shot results of Llama2-7B}
    \label{tab:performance}
\end{table}

\subsection{OOD and Ablation Experiment Results}

 Here, we supplement the experimental results of LoRA and AAG under supervised fine-tuning in closed-book settings and the ablation results of feedforward neural network (FFN) and Long Context Distillation (LCD). It can be observed that our method like LoRA, belongs to parameter-efficient fine-tuning, and because we share the Hypernetwork to generate LoRA adapter weights, we fine-tune fewer parameters. 
 
 From Table \ref{ffn}, it can be seen that releasing FFN can bring more performance improvement, possibly because adding LoRA in Attention cannot fully utilize enough knowledge \cite{KformerKI}. With the support of LCD, performance is further improved, with an average increase in EM of +5\%. This also proves the effectiveness of our proposed LCD. In comparison with AAG and LoRA, it becomes more evident that LoRA tends to transfer knowledge to the LoRA module, resulting in low generalization. Our method enhances knowledge activation through dynamic generation, showing significant effects not only ID but also in OOD.

\subsection{Error Analysis}
\label{error}
Using LLM as a knowledge base inevitably leads to hallucinations, which is a significant area of research in LLM development. In our quality analysis, we sampled 100 generated documents. As shown in Table \ref{tab:hallucinations_meaningless}, we found that hallucinations occurred with a probability of 4\%, while the occurrence of meaningless text, such as repeated values, was 6\%. Consequently, the impact of hallucinations in our method is relatively minor.

\subsection{Number of Document Compression}

In the first stage, we sampled 30,000 instances from the training sets of NQ and TQA, respectively, and used all 3,417 instances from the WebQ training set. To determine the number of retrieved documents to use for each dataset in stage 1, we conducted tests using the FiD (T5-Base) experiment. As shown in Table \ref{tab:performance_document_compression}, we can find that compressing five documents yielded relatively good performance. Consequently, we decided to compress five documents for each instance.

\begin{table}[htbp]
\centering
\resizebox{\linewidth}{!}{
\begin{tabular}{lccccc}
\hline
\textbf{} & \textbf{1} & \textbf{5} & \textbf{10} & \textbf{20} & \textbf{50} \\
\hline
\textbf{FiD} & 34.69 & 41.27 & - & - & 46.59 \\
\hline
\multicolumn{6}{l}{\textit{\textbf{\# Document Compression}}} \\

\textbf{AAG} & 32.57 & 38.19 & 35.17 & 32.12 & 36.83 \\
\hline
\end{tabular}
}
\caption{Performance Metrics for Different Configurations}
\label{tab:performance_document_compression}
\end{table}

\section{Case Study}
\label{case}

This study illustrates the differences in how three paradigms—RAG, GAG, and AAG—utilize documents during reasoning as shown in Table \ref{tab:gsm}. RAG retrieves ten documents from an external knowledge base, while GAG employs ChatGPT to generate ten documents with higher similarity. For illustration, we present only the content of the first document. Conversely, AAG uses its proprietary context generator to create virtual compressed documents containing more information. The token counts in parentheses show that AAG requires significantly fewer tokens for processing compared to the other two methods, thus enhancing inference speed and reducing computational overhead.

\begin{table*}[htbp]
\centering
\renewcommand\arraystretch{1.15}
\resizebox{\linewidth}{!}{
\begin{tabular}{lcp{16cm}}
\hline
\multicolumn{3}{l}{\makecell[l]{Question: what style of art did henri matisse use?}}
\\
\hline
\textbf{Method} & \textbf{\# Docs} & \textbf{Documents} \\
\hline
\textbf{RAG} & 10 & Henri Matisse Henri Émile Benoît Matisse (; 31 December 1869 – 3 November 1954) was a French artist, known for both his use of colour and his fluid and original draughtsmanship. He was a draughtsman, printmaker, and sculptor, but is known primarily as a painter. Matisse is commonly regarded, along with Pablo Picasso, as one of the artists who best helped to define the revolutionary developments in the visual arts throughout the opening decades of the twentieth century, responsible for significant developments in painting and sculpture. The intense colorism of the works he painted between 1900 and 1905 brought him... (\textcolor{red}{1860 tokens})  \\
\hline
\textbf{GAG} & 10 &Henri Matisse is considered one of the most important artists of the 20th century. He is known for his use of color and his distinctive style of painting and sculpture. Matisse was a member of the French avant-garde movement and his work was influenced by other artists such as Paul C\u00e9zanne and Pablo Picasso... (\textcolor{red}{1540 tokens})\\
\hline
\textbf{AAG}(ours) &1 & Henri Matisse context: a period of convalescence attack of appendic. He discovered \"a kind of paradise as he later described it, to become an deeply disappointing his. In 19 he returned to study art the Acad\u00e9mie and became a student of William-Adolphe Bouguereau Gustave Moreau Initially he painted still lif and landscapes a traditional style at which reasonable proficiency Mat was influenced the works earlier masters such as Jean-Bapt-Sim\u00e9on Ch Nicolas Pous Watteau, as well artists, such as \u00c9douard Manet  a body of work spanning over a half-century, won him recognition as a leading figure in modern art. Matisse was born in Le Cateau-Cambr\u00e9sis, in the Nord department in northern France, the oldest son of a prosperous grain merchant. He grew up in Bohain-en-Vermandois, Picardie, France. In 1887 he went to Paris to study law, working as a court administrator in Le Cateau-Cambr\u00e9sis after gaining his qualification. He first started to paint in 1889, after his mother brought him art supplies (\textcolor{green}{280 tokens})\\
\hline
\end{tabular}}
\caption{A inference case with used documents from WQ.}
\label{tab:gsm}
\end{table*}

\begin{table*}[htbp]
\centering
\resizebox{\linewidth}{!}{
\begin{tabular}{lC{3.3em}C{5.0em}C{2.5em}C{2.5em}C{2.5em}C{2.5em}C{2.5em}C{2.5em}C{2.5em}C{2.5em}C{2.5em}}
\hline
    \multirow{2}{*}{\bfseries Models} & {\bfseries \# Docu-} & {\bfseries Trainable} & \multicolumn{3}{c}{\bfseries NQ} & \multicolumn{3}{c}{\bfseries TQA} & \multicolumn{3}{c}{\bfseries WQ} \\ 
  & {\bfseries ments} & {\bfseries Params} & \underline{NQ} & TQA & WQ & NQ & \underline{TQA} & WQ & NQ & TQA & \underline{WQ} \\
\hline
    T5 & 0 & 220M & 22.16 & 3.18 & 4.12 & 2.65 & 21.8 & 3.15 & 0.88 & 2.95 & 28.3\\
    LoRA-Base & 0 & 28.3M & 5.43 & 3.15 & 4.02 & 0.00 & 9.60 & 0.00 & 0.22 & 1.77 & 20.47 \\
    ~~~~~~~w FFN & 0 & 141.5M & 16.17 & 4.71 & 6.89 & 3.15 & 21.16 & 0.00 & 1.33 & 3.04 & 26.38 \\
    ~~~~~~~w FFN \& LCD & 0 & 141.5M & 21.37 & 2.82 & 6.89 & 1.99 & 17.94 & 3.74 & 0.00 & 2.82 & 32.50 \\
    AAG & 0 & 26.1M & 5.31 & 3.82 & 5.71 & 0.22 & 10.34 & 2.12 & 0.55 & 2.30 & 16.58 \\
    ~~~~~~~w FFN & 0 & 139.3M & 21.05 & 4.52 & 6.50 & 3.51 & 19.08 & 3.15 & 2.11 & 3.84 & 28.17 \\
    ~~~~~~~w FFN \& LCD & 0 & 141.5M & 23.89 & 6.21 & 10.94 & 5.31 & 22.69 & 6.30 & 3.23 & 5.10 & 30.31 \\
\hline
    T5-l & 0 & 770M & 28.5* & 3.18 & 4.12 & 2.65 & 28.7* & 3.15 & 0.88 & 2.95 & 30.6*\\
    LoRA-l & 0 & 42.5M & 4.42 & 6.50 & 7.87 & 3.98 & 10.03 & 3.94 & 1.99 & 6.71 & 18.11 \\
    ~~~~~~~w FFN & 0 & 445.1M & 17.70 & 7.49 & 8.66 & 3.54 & 23.87 & 4.72 & 0.00 & 5.65 & 29.13 \\
    ~~~~~~~w FFN \& LCD & 0 & 445.1M & 28.32 & 4.52 & 10.94 & 5.31 & 25.71 & 6.12 & 1.75 & 4.52 & 29.92 \\
    AAG-l & 0 & 34.8M & 7.08 & 8.90 & 9.45 & 4.42 & 13.14 & 8.66 & 2.43 & 10.17 & 17.72 \\
    ~~~~~~~w FFN & 0 & 437.5M & 23.01 & 8.33 & 11.02 & 3.51 & 20.08 & 3.15 & 3.51 & 5.65 & 31.50 \\
    ~~~~~~~w FFN \& LCD & 0 & 437.5M & 29.32 & 10.17 & 14.06 & 7.02 & 30.11 & 7.81 & 2.65 & 7.06 & 32.68 \\
\hline
\end{tabular}
}
\caption{OOD and ablation experiment results in closed-book setting. * denotes the results are from the existing papers and LCD denotes Long Context Distillation.}
\label{ffn}
\end{table*}

\begin{table*}[htbp]
    \centering
    \resizebox{\linewidth}{!}{
    \begin{tabular}{L{10cm}|l}
        \toprule
        \textbf{Hallucinations} & \textbf{Meaningless} \\
        \midrule
        4\% & 6\% \\
        \midrule
        Question: When is the next Deadpool movie being released? & Question: Who got the first Nobel Prize in Physics? \\
        \hline
        Document: "\textcolor{purple}{Deadpool (film) Deadpool is a 2016 American superhero film} based on the Marvel Comics character of the same name, produced by Marvel Studios and distributed by Walt Disney Studios Motion Pictures. & Document: \textcolor{purple}{The Nobel Prize is not a prize in itself.} \\
        \hline
        Correct answer: May 18, 2018 & Correct answer: Wilhelm Conrad Röntgen \\
        \bottomrule
    \end{tabular}
    }
    \caption{Hallucinations and Meaningless Analysis.}
    \label{tab:hallucinations_meaningless}
\end{table*}

\begin{table*}[htbp]
\centering
% \vspace{-0.5em}
\renewcommand\arraystretch{1.2}
\resizebox{\linewidth}{!}{
\begin{tabular}{ll}
    \textbf{Methods} & \textbf{Prompt} \\ \hline
    \textbf{CBQA} & \makecell[l]{Please write a high-quality answer for the given question using your knowledge. \\Only give me the answer and do not output any other words.\\Question: \{question\}\\Answer: } \\ \hline
     \textbf{Retrieval} & \makecell[l]{Please write a high-quality answer for the given question using only the provided\\search results (some of which might be irrelevant). Only give me the answer\\and do not output any other words.\\Context: \{context\}\\Answer the question based on the given passages.\\Question: \{question\}\\Answer:} \\ \hline
      \textbf{Awakening} & \makecell[l]{Please write a high-quality answer for the given question using your knowledge \\ and the provided imagined compressed results (some of which might be irrelevant).\\ Only give me the answer and do not output any other words.\\Generated Context: \{context\}\\Answer the question based on your knowledge and the given generated context.\\Question: \{question\}\\Answer:} \\ \hline
\end{tabular}
}
\caption{Prompts for different methods on Zero-Shot setting. \textbf{CBQA} denotes closed-book QA that just prompts the model with the question.}
% \vspace{-2.2em}
\label{prompts}
\end{table*}

\begin{table*}[htbp]
\centering
% \vspace{-0.5em}
\renewcommand\arraystretch{1.05}
\resizebox{\linewidth}{!}{
\begin{tabular}{lC{5em}C{3.5em}C{3em}C{4.5em}C{3.5em}}
\hline
% \multirow{2}{*}{Name} & \multicolumn{2}{c|}{FAAG}  \\ \cline{2-3}
% \multirow{2}{*}{Class} & \multirow{2}{*}{Method} & \multicolumn{5}{c}{UMLS} & \multicolumn{5}{c}{FB15K-237} \\
%  & & MR & MRR & Hits@1 & Hits@1 & Hits@1 & MR & MRR & Hits@1 & Hits@1 & Hits@1
% & \multicolumn{5}{c}{FB15K-237} & \multicolumn{5}{c}{WN18RR} 
    \textbf{Models} & \textbf{Reader Params} & \textbf{\# Documents} & \textbf{NQ} & \textbf{TriviaQA} & \textbf{WebQ} \\ \hline
    \textit{\# Closed-book Setting} \\
    T5$^*$ \cite{T5} & 220M & 0 & 25.9 & 23.8 & 27.9 \\
    T5-l$^*$ \cite{T5} & 770M & 0 & 28.5 & 28.7 & 30.6 \\
    T5-xl \cite{T5} & 3b& 0 & 28.30 & 33.92 & 34.43 \\
    LoRA-Base & 220M & 0 & 5.43 & 9.60 & 20.47 \\
    LoRA-l  & 770M & 0 & 17.70 & 23.87 & 29.13 \\
    LoRA-xl & 3b & 0 & 23.15 & 32.16 & 35.24 \\
    AAG (Ours) & 220M & 0 & 23.89 & 22.69 & 30.31 \\
    AAG-l (Ours) & 770M & 0 & 29.32 & 30.11 & 32.68 \\
    AAG-xl (Ours) & 3b & 0 & \textbf{29.59} & \textbf{35.71} & \textbf{37.40} \\
    \hline
    \textit{\# Retrieval Augmented Generation} \\
    DPR$^*$ \cite{DPR} & 110M & 100 & 41.5 & 56.8 & 41.1 \\
    RAG$^*$ \cite{RAG} & 400M & 10 & 44.5 & 56.1 & 45.2 \\
    % FiD-xl & 3b & 20 & \textbf{55.18} & \textbf{72.92} & \textbf{52.85} \\
    FiD$^*$ \cite{fid} & 220M & 100 & 48.2 & 65.0 & 46.71 \\
    FiD-l$^*$ \cite{fid} & 770M & 100 & 51.4 & 67.6 & 50.52 \\
    FiD-xl \cite{fid} & 3b & 20 & \textbf{55.18} & \textbf{72.92} & \textbf{52.85} \\
    FiD-l$^*$  \cite{fid}& 770M & 10 & 46.7 & 61.9 & 48.1  \\
    FiD-xl$^*$ \cite{fid} & 3b & 10 & 50.1 & 66.3 & 50.8 \\
    EAR-l \cite{ear} &	770M	& 10	& 39.6	& 60.0	& - \\
    EAR-xl$^*$ \cite{ear} &	3b	& 10	& 42.3& 	64.6 & 	-\\
    RFiD-l \cite{rfid} & 770M & 10 & 48.3 & 63.4 & - \\
    RFiD-xl \cite{rfid} & 3b & 10 & 50.5 &  67.8 & - \\
    FILCO-xl$^*$ \cite{filco} &	3b	& 1	& 44.7	& 59.0	& - \\
    AAG (Ours) & 220M & 10 & 47.01 & 64.95 & 46.36  \\
    AAG-l (Ours) & 770M & 10 & 49.92 & 69.67 & 51.52 \\
    AAG-xl (Ours) & 3b & 5$^\ddagger$ & 50.87 & 70.34 & \underline{52.78}  \\
    AAG-l (Ours) & 770M & 30 & \underline{53.1} & \underline{70.5} & 52.0 \\
    \hline
    \textit{\# Generation Augmented Generation} \\
    GENREAD-l (sampling)$^*$ \cite{generate} & 770M & 10$^\dagger$ & 40.3 & 67.8 & 51.5 \\
    GENREAD-l (clustering)$^*$ \cite{generate} &	770M	& 10$^\dagger$	& 43.5	& 70.2	& 53.5 \\
    GENREAD-xl (sampling)$^*$ \cite{generate} & 3b & 10$^\dagger$ & 42.6 & 69.6 & 52.6 \\
    GENREAD-xl (clustering)$^*$ \cite{generate} &	3b	& 10$^\dagger$	& 45.6	& \underline{71.6}	& 54.4 \\
    AAG (Ours) & 220M & 10$^\dagger$ & 46.22 & 66.70 & 51.43  \\
    AAG-l (Ours) & 770M & 10$^\dagger$ & \underline{48.83} & 70.85 & \underline{54.52} \\
    AAG-xl (Ours) & 3b & 5$^\dagger$$^\ddagger$ & \textbf{49.23} & \textbf{72.18} & \textbf{55.39}  \\
    \hline
    \textit{\# Awakening Augmented Generation (Ours)} \\
    LoRA-Base & 220M & 1$^\dagger$ & 34.51 & 54.05 & 32.28 \\
    LoRA-l & 770M & 1$^\dagger$ & 40.05 & 62.81 & 43.70 \\
    LoRA-xl & 3b & 1$^\dagger$ & 44.15 & 66.92 & 48.23 \\
    AAG & 220M & 1$^\dagger$ & 40.14 & 60.75 & 41.73 \\
    AAG-l & 770M & 1$^\dagger$ & 42.32 & 65.48 & 45.28 \\
    AAG-xl & 3b & 1$^\dagger$ & \textbf{46.51} & \textbf{68.38} & \textbf{50.45} \\
    \hline
\end{tabular}
}
\caption{Full QA performances (\%) of different methods on three datasets. The first part (closed-book setting) indicates that explicit documentation was not utilized; The latter three parts utilize explicit augmented documents. The best results are in bold. * means that those results are from existing papers, $^\dagger$ denotes that the number of documents is generated ($\ddagger$ indicates that the number of documents is reduced due to insufficient memory for distillation).}
% \vspace{-2.2em}
\label{fullsup}
\end{table*}

\end{document}